\documentclass[letterpaper, 10 pt, journal, twoside]{IEEEtran}  % Comment this line out if you need a4paper

\IEEEoverridecommandlockouts                              		% This command is only needed if 
\usepackage{graphics} % for pdf, bitmapped graphics files
\usepackage{graphicx}
\usepackage{epsfig}   % for postscript graphics files
\usepackage{amsmath}  % assumes amsmath package installed
\usepackage{amssymb}  % assumes amsmath package installed
\usepackage{eufrak}
\usepackage{cite}
\usepackage{multirow}
\usepackage{multicol}
\usepackage[table]{xcolor}
\usepackage{caption}
\usepackage{acronym}
\usepackage{subcaption}
\usepackage{soul}
\usepackage{titlesec}

\titlespacing\section{0pt}{8pt plus 2pt minus 2pt}{5pt plus 2pt minus 2pt}
\titlespacing\subsection{0pt}{8pt minus 2pt}{5pt minus 2pt}

\def\Vec#1{\!\!\hbox{$#1$\kern-0.38em\lower0.85em\hbox{$\vec{}\,$}}\,}%
\newcommand{\bbm}{\begin{bmatrix}}
\newcommand{\ebm}{\end{bmatrix}}
\DeclareMathAlphabet{\mbf}{OT1}{ptm}{b}{n}

\acrodef{VO}{visual odometry} 
\acrodef{VTR}[VT\&R]{Visual Teach and Repeat} 
\acrodef{RANSAC}{Random Sample Consensus}    
\acrodef{UTIAS}{University of Toronto Institute for Aerospace Studies}  
\acrodef{SURF}{Speeded-Up Robust Features}
\acrodef{RMSE}{root mean squared error}
\acrodef{DOF}{degrees of freedom}

\author{Mona Gridseth$^{1}$ and Timothy D. Barfoot$^{1}$%
\thanks{Manuscript received: September, 8, 2021; Revised December, 3, 2021; Accepted December, 13, 2021.}%Use only for final RAL version
\thanks{This paper was recommended for publication by Editor Sven Behnke upon evaluation of the Associate Editor and Reviewers' comments.
This work was generously supported by Clearpath Robotics and the Natural Sciences and Engineering Research Council of Canada (NSERC). We thank Yuchen Wu, Ben Congram, and Sherry Chen for their help with VT\&R.} %Use only for final RAL version
\thanks{$^{1}$Mona Gridseth and Timothy D. Barfoot are with the University of Toronto Institute for Aerospace Studies, University of Toronto, Canada  \tt\footnotesize{mona.gridseth@robotics.utias.utoronto.ca., tim.barfoot@utoronto.ca }}%
\thanks{Digital Object Identifier (DOI): see top of this page.}
}

\title{Keeping an Eye on Things: Deep Learned Features for Long-Term Visual Localization}

\begin{document}

\maketitle
%\thispagestyle{empty}
%\pagestyle{empty}

%%%%%%%%%%%%%%%%%%%%%%%%%%%%%%%%%%%%%%%%%%%%%%%%%%%%%%%%%%%%%%%%%%%%%%%%%%%%%%%%
\begin{abstract}
In this paper, we learn visual features that we use to first build a map and then localize a robot driving autonomously across a full day of lighting change, including in the dark. We train a neural network to predict sparse keypoints with associated descriptors and scores that can be used together with a classical pose estimator for localization. Our training pipeline includes a differentiable pose estimator such that training can be supervised with ground truth poses from data collected earlier, in our case from 2016 and 2017 gathered with multi-experience \ac{VTR}. We insert the learned features into the existing \ac{VTR} pipeline to perform closed-loop path following in unstructured outdoor environments. We show successful path following across all lighting conditions despite the robot's map being constructed using daylight conditions. Moreover, we explore generalizability of the features by driving the robot across all lighting conditions in new areas not present in the feature training dataset. In all, we validated our approach with 35.5 km of autonomous path following experiments in challenging conditions.  
\end{abstract}

\begin{IEEEkeywords}
Localization, Deep Learning for Visual Perception, Vision-Based Navigation
\end{IEEEkeywords}

%%%%%%%%%%%%%%%%%%%%%%%%%%%%%%%%%%%%%%%%%%%%%%%%%%%%%%%%%%%%%%%%%%%%%%%%%%%%%%%%

\section{Introduction}

\IEEEPARstart{L}{ong-term} navigation across large appearance change is a challenge for robots that use cameras for sensing. Visual Teach and Repeat (\ac{VTR}) tackles this problem with the help of multi-experience localization \cite{Paton2016}. While the user manually drives the robot to teach a path, \ac{VTR} builds a visual map. Afterwards, it localizes live images to the map, allowing the robot to repeat the taught path autonomously. As the environment changes, \ac{VTR} stores data from each path repetition (experience). These intermediate experiences are used to bridge the appearance gap as localizing to the initial map becomes more challenging. With the help of machine learning, we aim to remove the need for such intermediate experiences for localization. Previously, we trained a neural network that would predict relative pose for localization directly from two input images \cite{gridseth2020}. The network was able to directly predict pose across large appearance change without using intermediate experiences. However, it learned the full pose estimation pipeline, including the parts that are easily solved with classical methods. Moreover, it did not generalize well to new paths not seen in the training data. These results align with Sattler et al. \cite{Sattler2019}, who found that accuracy can be an issue for methods that regress global pose directly from sensor data. 

\begin{figure}
  \centering
  \includegraphics[width=0.4\textwidth]{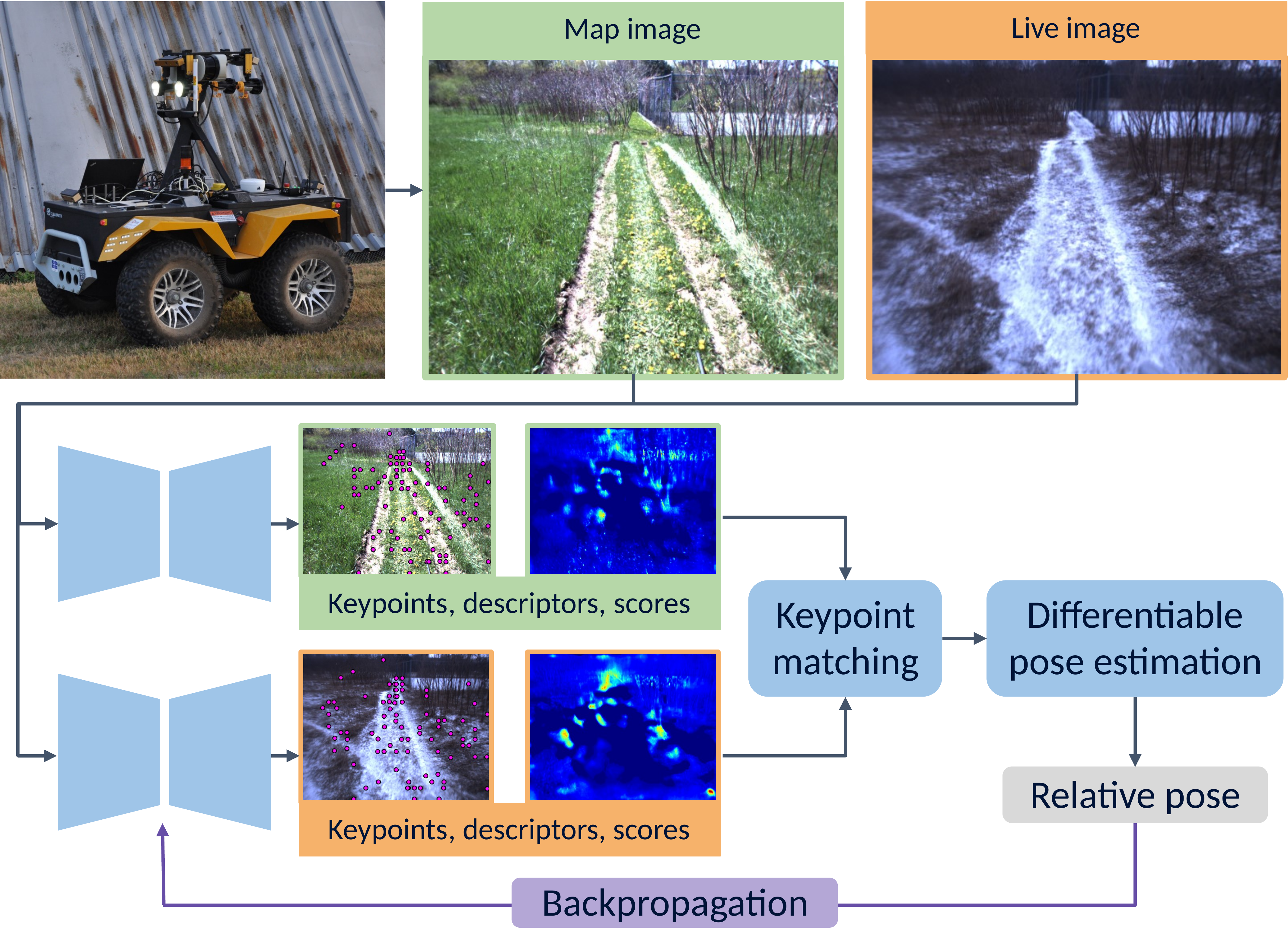}
  \caption{\small We train a neural network to predict keypoints, descriptors, and scores that can be used in classical pose estimation. We train on outdoor data collected with a Grizzly ground robot and later drive the same robot autonomously with the learned features despite severe lighting change.}
  \label{fig:overview}
  \vspace{-0.4cm}
\end{figure}
 
In this paper, we choose a different strategy. Instead of predicting poses directly from images, we use learning for the perception front-end of long-term localization, while the remaining components of pose estimation are implemented with classical tools, see Figure \ref{fig:overview}. We insert the learned features into the existing \ac{VTR} pipeline and perform autonomous path following outdoors without needing intermediate bridging experiences. In particular, we teach a path (and build a visual map) at day time and repeat it for a full day including after dark. In another experiment, we show the learned features can generalize by driving in new areas not present in the feature training dataset. 
 
For training, we use data collected across lighting and seasonal change in 2016 and 2017 by a robot using multi-experience \ac{VTR}. We train a network to provide keypoints with associated descriptors and scores. Using a differentiable pose estimator allows us to backpropagate losses based on poses from the training data. The network is small enough to fit on a laptop GPU, fast enough to run in real time, and uses only two losses generated from pose ground truth. 

Our method differs from others that use learned features for localization across environmental change, such as \cite{piasco2019, Stumberg2020a, Stumberg2020b, Kasper2020, spencer2020,Sarlin2021}, since they test localization standalone, while we include our features in the full \ac{VTR} system. Gladkova et al. \cite{Gladkova2021} combine learned features for localization with \ac{VO}, but only test on datasets. Since our goal is to learn features for the path-following task, we focus on closed-loop performance. Good localization performance on datasets does not guarantee successful real-life autonomous driving, which involves interaction of localization with VO and path-tracking control. Sun et al. \cite{Sun2021} recently published closed-loop path following with learned features, but their lighting-change experiments are less extensive than ours.

\begin{figure*}[h]
  \centering
  \includegraphics[width=1.0\textwidth]{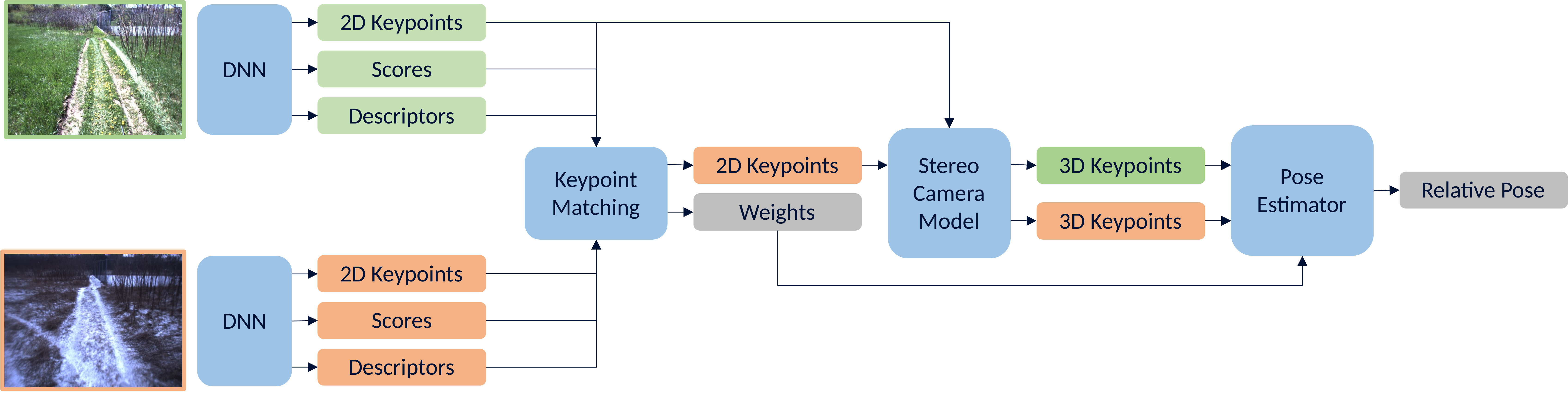}
  \caption{\small In the learning pipeline we pass images from a source frame (green) and a target frame (orange) to a neural network that predicts keypoints, descriptors, and scores. We match keypoints between the image frames, use the stereo camera model to find their corresponding 3D coordinates, and use the matched point pairs to estimate the relative pose.}
  \label{fig:pipeline}
  \vspace{-0.5cm}
\end{figure*}

\section{Related Work}

There has been a wide range of work on deep learning for pose estimation. Chen et al. \cite{chen2020survey} provide a thorough survey of deep learning for mapping and localization. Some research has focused on learning pose for localization directly from images via absolute pose regression \cite{Kendall2015}, relative pose regression \cite{laskar2017}, or combining learning for localization and \ac{VO} \cite{Valada2018}. Sattler et al. \cite{Sattler2019} note that learning pose directly from image data can struggle with accuracy.

More structure can be imposed on the learning problem by using features to tackle front-end visual matching, while retaining a classical method for pose estimation. A wide range of papers have been published on learning sparse visual features with examples in \cite{rosten2008, yi2016, ono2018, detone2018a, dusmanu2019, revaud2019, luo2020, sarlin2020, Wang2020}. Another option is to learn descriptors densely for the image \cite{Lv2019, Tang2019, Stumberg2020a, Stumberg2020b, Kasper2020, Xu2020}, which can also be used for sparse matching \cite{spencer2020, Sarlin2021}.

Although several papers have tackled descriptor learning for challenging appearance change, including seasonal change,\cite{piasco2019, Stumberg2020a, Stumberg2020b, Kasper2020, spencer2020, Sarlin2021}, they test localization standalone. In our work we include the learned features in the \ac{VTR} pipeline, where relative localization to a map is combined with \ac{VO} for long-term path following. Moreover, we use mostly off-road data with fewer permanent structures, where appearance change can be more drastic than in urban areas.

Sarlin et al. \cite{Sarlin2021} show good generalization to new domains with their learned features. For instance, they are able to use features trained on outdoor data for indoor localization. In \cite{Stumberg2020a, Stumberg2020b, spencer2020}, the authors generalize to unseen seasonal conditions on the same path, while Piasco et al. \cite{piasco2019} train and test on different sections of a path. In our work, we drive three paths that are partially or entirely in areas not included in the training data, showing the generalizability of our features to novel environments.

The work from Gladkova et al. \cite{Gladkova2021} is close to ours as they integrate learned features for localization in a \ac{VO} pipeline. The localization poses are used as a prior for front-end tracking and integrated into back-end bundle adjustment. Global localization poses are fused with \ac{VO} estimates. While the authors test on urban datasets with seasonal change, we test our approach in closed loop on a robot. Sun et al. \cite{Sun2021} recently published work, where they drive a robot in closed loop using learned features in VT\&R. They test day-to-night localization, though their experiments are less extensive with a shorter time range. Furthermore, they test lighting change in on-road areas such as a parking lot and by a church, while we also include more challenging off-road driving. 

We base our learning pipeline and network architecture on the design presented by Barnes and Posner \cite{barnes2020}, which learns keypoints, descriptors, and scores for radar localization. We chose this method because it required only a pose loss for supervision and provided a simple network design with a fully differentiable learning pipeline. In particular, a fixed number of keypoints is detected across the image removing any need for pruning with techniques such as non-maximal suppression (NMS). Overall, network size and run-time is important for our real-time application. We made attempts at unsupervised alternatives, similar to \cite{Tang2020}, but struggled with estimation in the longitudinal direction and therefore opted for a supervised approach. 
  
\section{Methodology}
\label{sec:methodology}

Our fully differentiable training pipeline takes a pair of source and target stereo images and estimates the relative pose, $\mbf{T}_{ts} \in SE(3)$, between their associated frames. We build on the approach for radar localization in \cite{barnes2020} with some modifications to use a vision sensor. In short, the neural network detects keypoints and computes their descriptors and scores. We match keypoints from the source and target before computing their 3D coordinates with a stereo camera model. Finally, the point correspondences are used in a differentiable pose estimator. For an overview, see Figure \ref{fig:pipeline}.

\subsection{Keypoint Detection and Description}

\begin{figure}
  \centering
  \includegraphics[width=0.46\textwidth]{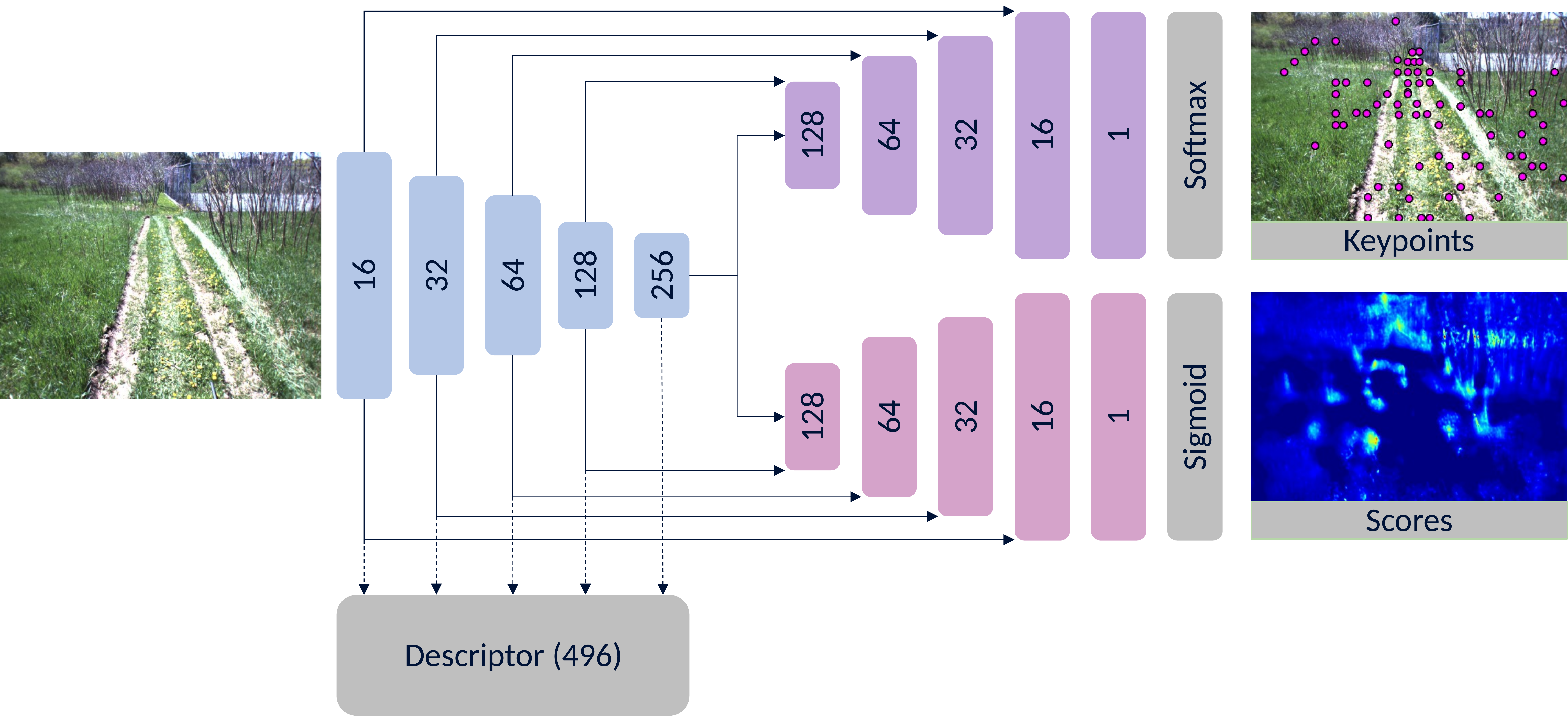}
  \caption{\small Neural network architecture with one decoder to detect keypoints and one to learn scores for all pixels. Descriptors for all pixels are computed by resizing and concatenating the output feature maps of each encoder layer.}
  \label{fig:network}
  \vspace{-0.5cm}
\end{figure}

We start by detecting sparse 2D keypoints, $\mbf{q} = \bbm u_\ell & v_\ell \ebm^T$, at sub-pixel locations in the left stereo image and computing descriptor vectors, $\mbf{d} \in \mathbb{R}^{496}$, and scores, $s \in [0, 1]$, for all pixels in the image. The descriptor and score for a given keypoint is found using bilinear interpolation. The score determines how important a point is for pose estimation. We pass an image to an encoder-decoder neural network following the design from \cite{barnes2020}, illustrated in Figure \ref{fig:network}. After the bottleneck, the network branches into two decoder branches, one to compute keypoint coordinates and one for the scores. We divide the image into size $16\times16$ windows and detect a keypoint for each one by taking the weighted average of the pixel coordinates. We get the weights by computing the softmax over the network output for each window. Applying a sigmoid to the output of the second decoder branch gives us the scores. Finally, the descriptors are found by resizing and concatenating the feature maps of each encoder block, leaving us with a length 496 descriptor vector for each pixel.

\subsection{Keypoint Matching}
\label{sec:matching}

We have a set of $N$ keypoints for the source image and we need to perform data association between these and points in the target image. Descriptors are compared using zero-normalized cross correlation (ZNCC), which for real matrices is the same as cosine distance and the resulting value will be in the range $[-1, 1]$. For each keypoint in the source image, $\mbf{q}_{s}^i, i \in [1, N]$, we compute a matched point, $\hat{\mbf{q}}_{t}^i$, in the target image. This point is the weighted sum of all image coordinates in the target image, where the weight is based on how well descriptors match: 
\begin{equation}
\label{eq:matching}
\hat{\mbf{q}}_{t}^i = \sum_{j=1}^{M} \sigma(\tau f_{\rm zncc}(\mbf{d}_{s}^i, \mbf{d}_{t}^j))\mbf{q}_{t}^j.
\end{equation}
$M$ is the total number of pixels in the target image, $f_{\rm zncc}(\cdot)$ computes the ZNCC between the descriptors, and $\sigma(\cdot)$ takes the temperature-weighted softmax with $\tau$ as the temperature, which is determined empirically by trying a range of values. The keypoint matching is differentiable. We found that using all target pixels worked better in practice than only including keypoints detected in the target image. Finally, we find the descriptor, $\hat{\mbf{d}}_t^i$, and score, $\hat{s}_t^i$, for each computed target point using bilinear interpolation. %An example of detected and matched keypoints and corresponding scores for a pair of images can be see in Figure \ref{fig:feature_example}.   

\subsection{Stereo Camera Model}
\label{sec:stereo_model}

In order to estimate the pose from matched 2D keypoints, we need to get their corresponding 3D coordinates, which is straightforward with a stereo camera. The camera model, $\mbf{g}(\cdot)$, for a pre-calibrated stereo rig maps a 3D point, $\mbf{p} = \bbm x & y & z \ebm^T$, in the camera frame to a left stereo image coordinate, $\mbf{q}$, together with disparity, $d = u_\ell - u_r$, as follows:
\begin{equation} \label{eq:cam_model}
\mbf{y} = \bbm u_\ell \\ v_\ell \\ d \ebm = \bbm \mbf{q} \\ d \ebm = \mbf{g}(\mbf{p}) = \bbm f_u & 0 & c_u & 0 \\ 0 & f_v & c_v & 0 \\ 0 & 0 & 0 & f_ub \ebm \, \frac{1}{z} \bbm x \\ y \\ z \\ 1 \ebm, 
\end{equation} 
where $f_u$ and $f_v$ are the horizontal and vertical focal lengths in pixels, $c_u$ and $c_v$ are the the camera's horizontal and vertical optical centre coordinates in pixels, and $b$ is the baseline in metres. We use the inverse stereo camera model to get each keypoint's 3D coordinates:
\begin{equation} \label{eq:inv_cam_model}
\mbf{p} = \bbm x \\ y \\ z \ebm = \mbf{g}^{-1} (\mbf{y}) = \frac{b}{d} \bbm  u_\ell - c_u  \\ \frac{f_u}{f_v}\left( v_\ell - c_v \right) \\ f_u \ebm.
\end{equation}
We perform stereo matching to obtain disparity, $d$, by using \cite{Hirschmuller2008} as implemented in OpenCV.  

\subsection{Pose Estimation}

Given the correspondences between the source keypoints, $\mbf{q}_s^i$, and matched target keypoints, $\hat{\mbf{q}}_t^i$, we can compute the relative pose from the source to the target, $\mbf{T}_{ts} = \left[ \begin{matrix} \mbf{C}_{ts} & \mbf{r}^{st}_t \\ \mbf{0} & 1 \end{matrix} \right] $, where $\mbf{r}^{st}_t$ is the translation from the target frame to the source frame given in the target frame. As described in Section \ref{sec:stereo_model}, we use the inverse stereo camera model (\ref{eq:inv_cam_model}) to compute 3D coordinates, $\mbf{p}_s^i$ and $\hat{\mbf{p}}_t^i$, from the corresponding 2D keypoints. This allows us to minimize the following cost:
\begin{equation} \label{eq:pose_est}
J = \sum_{i=1}^{N} w^i||(\mbf{C}_{ts}\mbf{p}_s^i + \mbf{r}^{st}_t) - \hat{\mbf{p}}_t^i||^2_2.
\end{equation} 
The minimization is implemented using Singular Value Decomposition (SVD) (more details can be found in \cite{Barfoot2017}), which is differentiable. The weight, $w^i \in [0, 1]$, for a matched point pair is a combination of the learned point scores, $s_s^i$ and $\hat{s}_t^i$, and how well the descriptors match:
\begin{equation} \label{eq:weight_learned}
w^i = \frac{1}{2}(f_{\rm zncc}(\mbf{d}_{s}^i, \hat{\mbf{d}}_{t}^i) + 1)s_s^i\hat{s}_t^i.
\end{equation}
We additionally remove large outliers at training time based on ground truth keypoint coordinate error and using \ac{RANSAC} at inference.

\subsection{Loss Functions}
Barnes and Posner \cite{barnes2020} supervise training using only a loss on the estimated pose. We found this was insufficient for our approach and also include a loss on the 3D coordinates of the matched keypoints, similar to \cite{christiansen2019}. We generate the keypoint ground truth from the poses and do not require additional keypoint correspondence data. These losses are sufficient and we do not need to add losses for descriptors or scores. For our training datasets, we only use a subset of the pose \ac{DOF} for supervision due to accuracy variability in the remaining DOF for some sequences. Specifically, we use the robot longitudinal direction, $\alpha$, lateral offset, $\beta$, and heading, $\gamma$. For this reason, we modify the keypoint loss to only use these \ac{DOF}s.

Using (\ref{eq:inv_cam_model}), we can compute the 3D coordinates of the keypoints in the source and target camera frames. Given the ground truth pose, $\mbf{T}_{ts}$, we form a pose 
\begin{equation}
\mbf{T}_{ts}^{\prime} = \left[ \begin{matrix} \mbf{C}_{ts}^{\prime} & {\mbf{r}^{st}_t}^{\prime} \\ \mbf{0} & 1 \end{matrix} \right] = \left[ \begin{matrix} \cos(\gamma) & -\sin(\gamma) & 1 & \alpha \\ \sin(\gamma) & \cos(\gamma) & 1 & \beta  \\ 0 & 0 & 1 & 0  \\ 0 & 0 & 0 & 1 \end{matrix} \right],
\end{equation}
that we use to transform the source keypoints. Because we transform the source points in the plane, we only compare the $x$ and $y$ point coordinates: 
\begin{equation}
\mathcal{L}_{\rm keypoint} = \sum_{i=1}^N ||\mbf{T}_{ts}^{\prime}\left. \mbf{p}_s^i \right|_{z=0} - \left.\hat{\mbf{p}}_t^i\right|_{z=0} ||_2^2.
\end{equation}
Forming $\mbf{T}_{ts}^{\prime}$ from the ground truth pose and $\hat{\mbf{T}}_{ts}^{\prime}$ from the estimated pose, we get the following pose loss:
\begin{equation}            
\mathcal{L}_{\rm pose} = || {{}\hat{\mbf{r}}^{st}_t}^{\prime} - {\mbf{r}^{st}_t}^{\prime}||_2^2  + \lambda||\hat{\mbf{C}}_{ts}^{\prime}{\mbf{C}_{ts}^{\prime}}^T - \mbf{1}||_2^2,
\end{equation}
where $\lambda$ is used to balance rotation and translation and $\mbf{1}$ is the identity matrix. The total loss is a weighted sum of $\mathcal{L}_{\rm keypoint}$ and $\mathcal{L}_{\rm pose}$, where the weight is determined empirically to balance the influence of the two loss terms.

\begin{figure}
    \centering
    \begin{minipage}{0.24\textwidth}
        \centering
        \includegraphics[width=\textwidth]{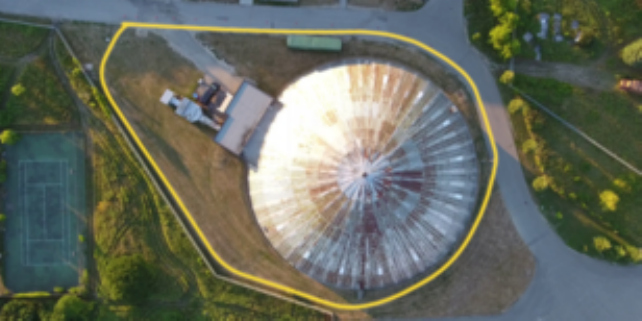}
    \end{minipage}%
    \hspace{0.10mm}
    \begin{minipage}{0.24\textwidth}
        \centering
        \includegraphics[width=\textwidth]{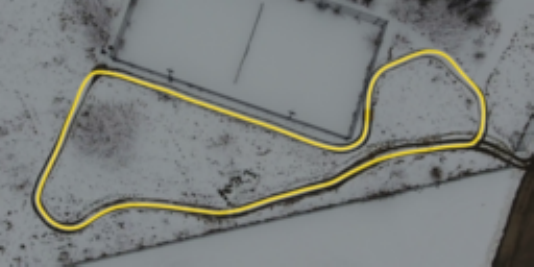}
    \end{minipage}%
    \caption{\small The paths for the In-The-Dark (left) and Multiseason (right) training datasets.}
    \label{fig:paths}
    \vspace{-0.4cm}
\end{figure}

\begin{figure}
    \centering
    \begin{subfigure}[]{0.225\textwidth}
  		\centering
        \includegraphics[width=\textwidth,trim={1cm 0 1.5cm 0},clip]{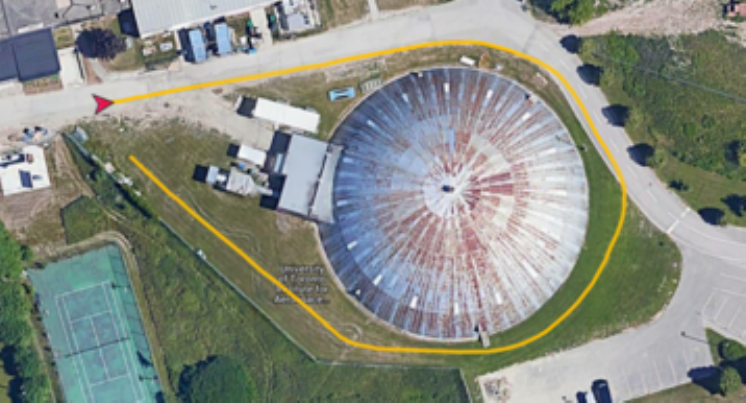}
    \end{subfigure}
    %\hspace{1.00mm}
    \begin{subfigure}[]{0.25\textwidth}
        \centering
        \includegraphics[width=\textwidth]{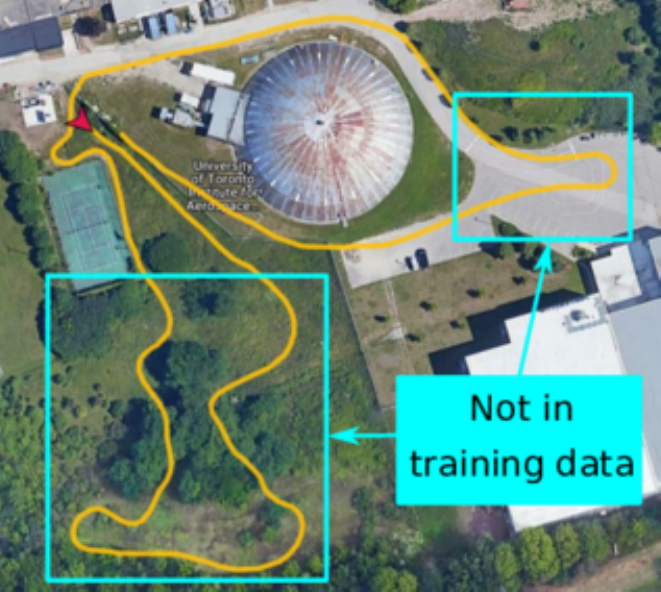}
    \end{subfigure}
    \par\smallskip
    \hspace{-27.5mm}
    \begin{subfigure}[]{0.11\textwidth}
        \centering
        \includegraphics[width=\textwidth,trim={4cm 0.5cm 3cm 1.5cm},clip, angle=90]{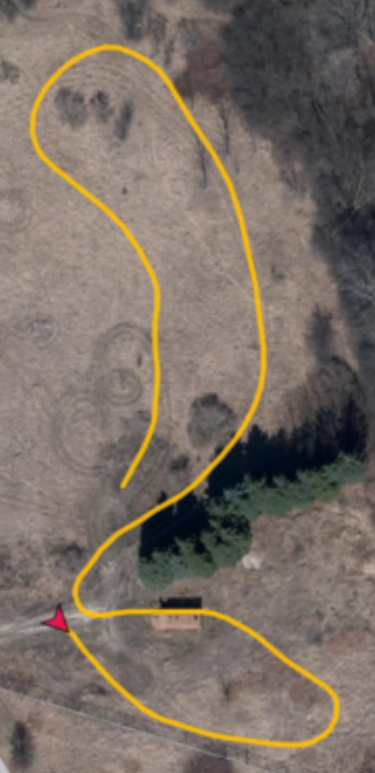}
    \end{subfigure}
    \hspace{20.0mm}
    \begin{subfigure}[]{0.11\textwidth}
        \centering
        \includegraphics[width=\textwidth,trim={4cm 1.0cm 4.03cm 2.0cm},clip, angle=90]{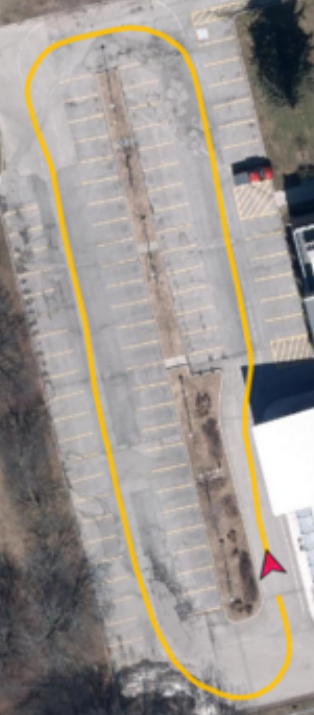}
    \end{subfigure}
    \caption{\small The paths taught with the robot for the lighting-change (top left) and first generalization (top right) experiments. For the generalization experiment the top left area of the path corresponds to the Multiseason path from Figure \ref{fig:paths}. The rectangles indicate the parts of the path that are outside of the training data. The second (field) and third (parking lot) generalization paths are on the bottom.}
    \label{fig:live_paths}
    \vspace{-0.5cm}
\end{figure}

\begin{figure*}
  \centering
  \begin{subfigure}[]{0.135\textwidth}
  		\centering
        \includegraphics[width=\textwidth]{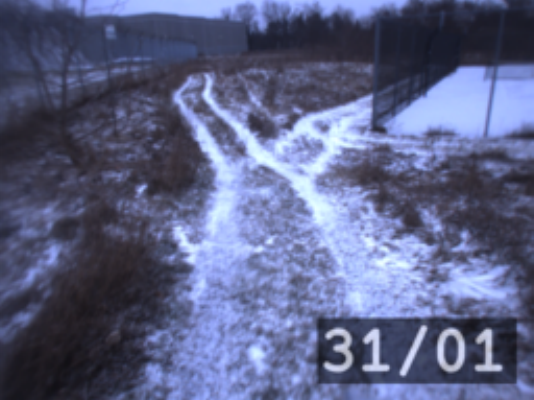}
  \end{subfigure}
  \begin{subfigure}[]{0.135\textwidth}
		\centering        
        \includegraphics[width=\textwidth]{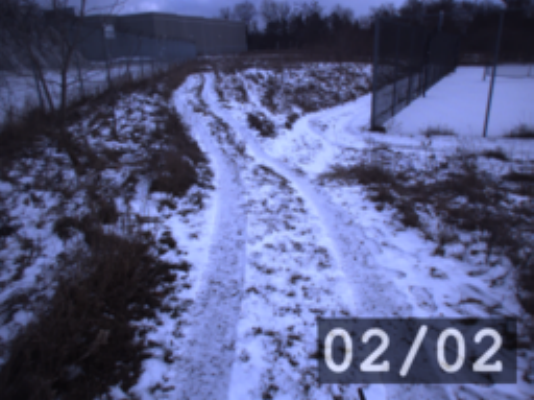}
  \end{subfigure}
  \begin{subfigure}[]{0.135\textwidth}
  		\centering
        \includegraphics[width=\textwidth]{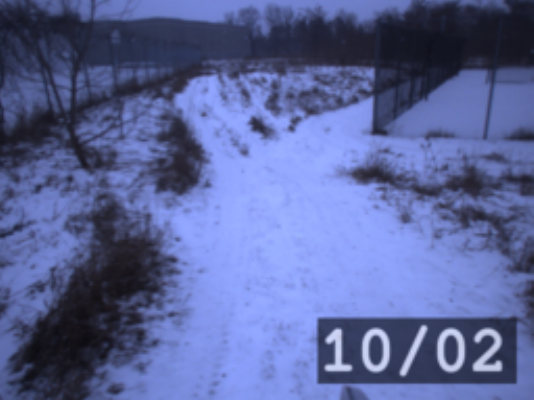}
  \end{subfigure}
  \begin{subfigure}[]{0.135\textwidth}
		\centering        
        \includegraphics[width=\textwidth]{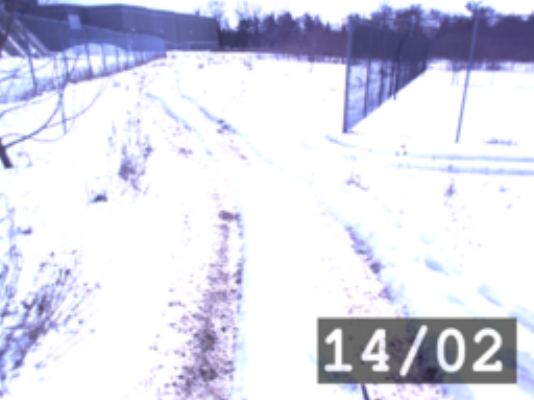}
  \end{subfigure}
  \begin{subfigure}[]{0.135\textwidth}
  		\centering
        \includegraphics[width=\textwidth]{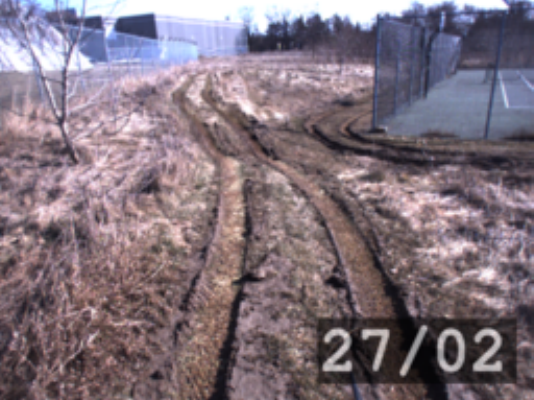}
  \end{subfigure}
    \begin{subfigure}[]{0.135\textwidth}
  		\centering
        \includegraphics[width=\textwidth]{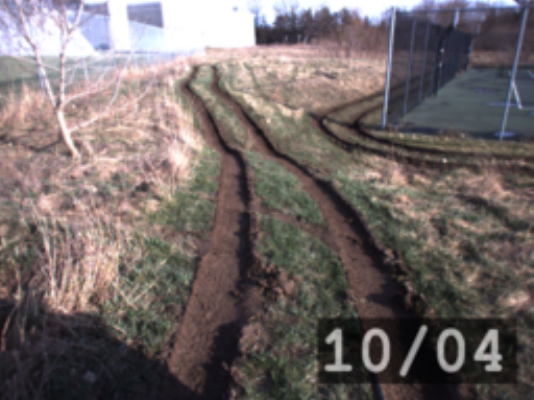}
  \end{subfigure}
  \begin{subfigure}[]{0.135\textwidth}
		\centering        
        \includegraphics[width=\textwidth]{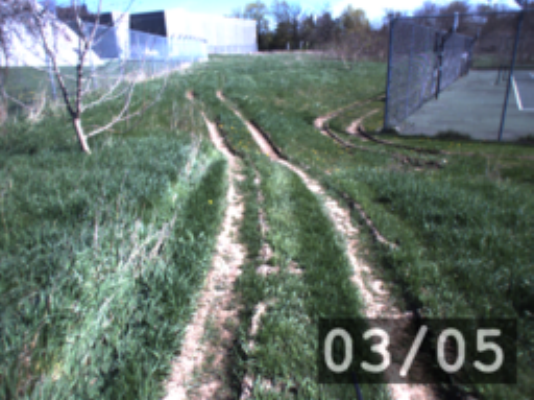}
  \end{subfigure}
  \caption{\small Images in the offline experiment taken by the robot camera from repeats labeled with the collection date (dd/mm).}
  \label{fig:offline_images}
  \vspace{-0.25cm}
\end{figure*}

\begin{figure*}
  \centering
  \begin{subfigure}[]{0.118\textwidth}
  		\centering
        \includegraphics[width=\textwidth]{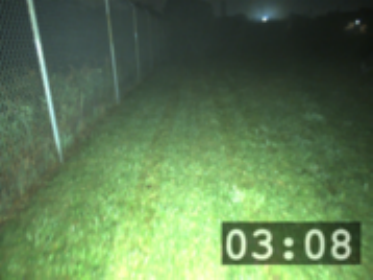}
  \end{subfigure}
  \begin{subfigure}[]{0.118\textwidth}
  		\centering
        \includegraphics[width=\textwidth]{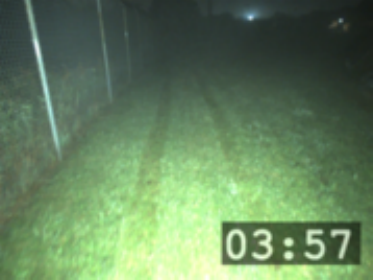}
  \end{subfigure}
  \begin{subfigure}[]{0.118\textwidth}
  		\centering
        \includegraphics[width=\textwidth]{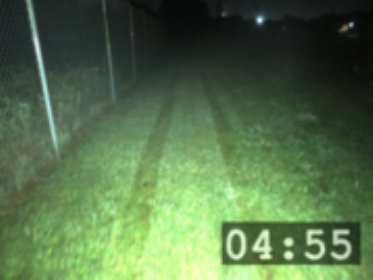}
  \end{subfigure}
  \begin{subfigure}[]{0.118\textwidth}
  		\centering
        \includegraphics[width=\textwidth]{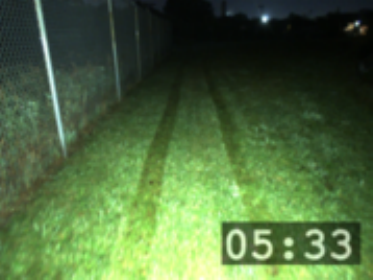}
  \end{subfigure}
  \begin{subfigure}[]{0.118\textwidth}
  		\centering
        \includegraphics[width=\textwidth]{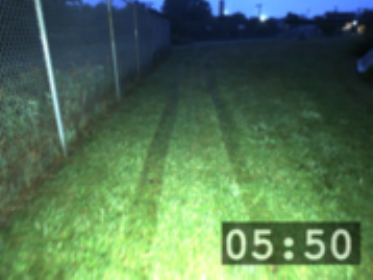}
  \end{subfigure}
  \begin{subfigure}[]{0.118\textwidth}
  		\centering
        \includegraphics[width=\textwidth]{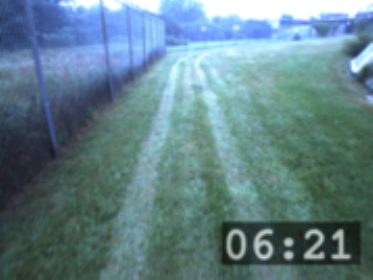}
  \end{subfigure}
  \begin{subfigure}[]{0.118\textwidth}
  		\centering
        \includegraphics[width=\textwidth]{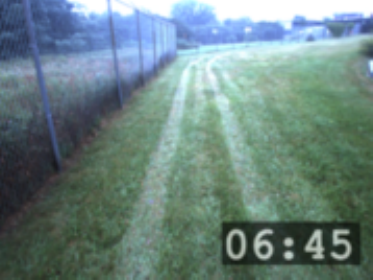}
  \end{subfigure}
  \begin{subfigure}[]{0.118\textwidth}
  		\centering
        \includegraphics[width=\textwidth]{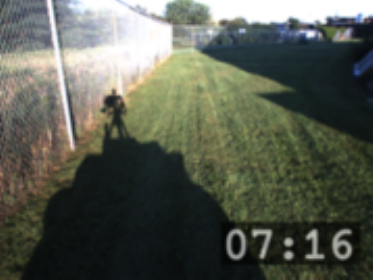}
  \end{subfigure}
  \par\smallskip
  \begin{subfigure}[]{0.118\textwidth}
  		\centering
        \includegraphics[width=\textwidth]{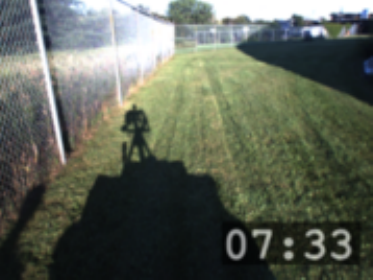}
  \end{subfigure}
  \begin{subfigure}[]{0.118\textwidth}
  		\centering
        \includegraphics[width=\textwidth]{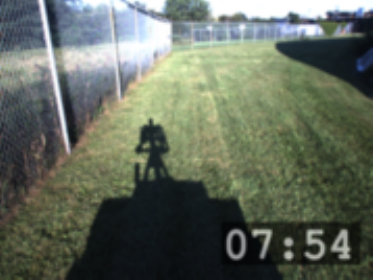}
  \end{subfigure}
  \begin{subfigure}[]{0.118\textwidth}
  		\centering
        \includegraphics[width=\textwidth]{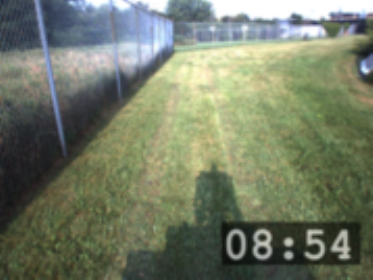}
  \end{subfigure}
  \begin{subfigure}[]{0.118\textwidth}
  		\centering
        \includegraphics[width=\textwidth]{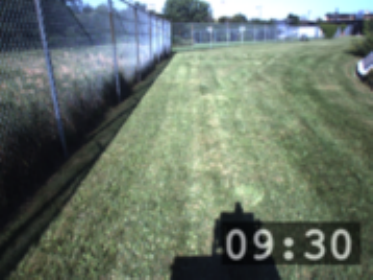}
  \end{subfigure}
  \begin{subfigure}[]{0.118\textwidth}
  		\centering
        \includegraphics[width=\textwidth]{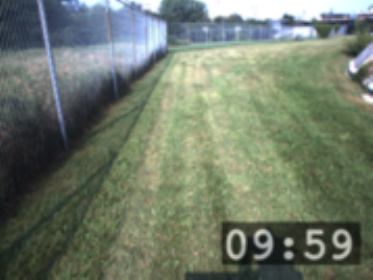}
  \end{subfigure}
  \begin{subfigure}[]{0.118\textwidth}
  		\centering
        \includegraphics[width=\textwidth]{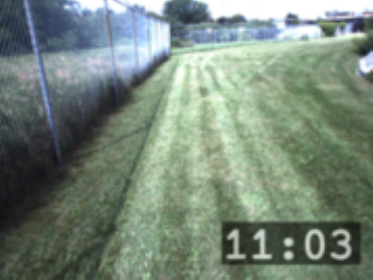}
  \end{subfigure}
  \begin{subfigure}[]{0.118\textwidth}
  		\centering
        \includegraphics[width=\textwidth]{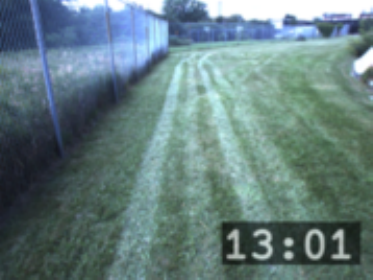}
  \end{subfigure}
  \begin{subfigure}[]{0.118\textwidth}
  		\centering
        \includegraphics[width=\textwidth]{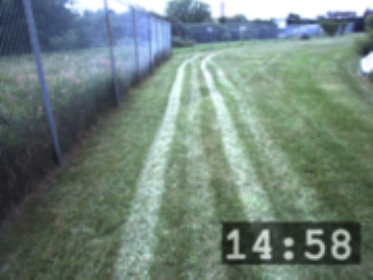}
  \end{subfigure}
  \par\smallskip
  \begin{subfigure}[]{0.118\textwidth}
  		\centering
        \includegraphics[width=\textwidth]{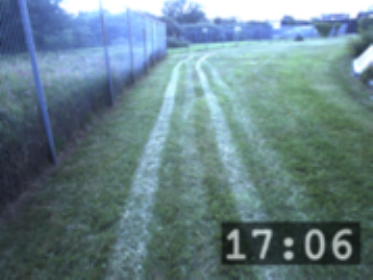}
  \end{subfigure}
  \begin{subfigure}[]{0.118\textwidth}
  		\centering
        \includegraphics[width=\textwidth]{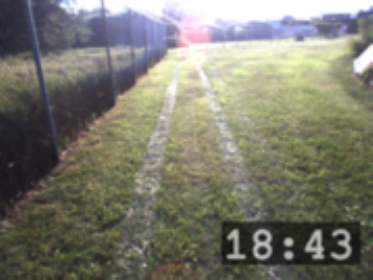}
  \end{subfigure}
  \begin{subfigure}[]{0.118\textwidth}
  		\centering
        \includegraphics[width=\textwidth]{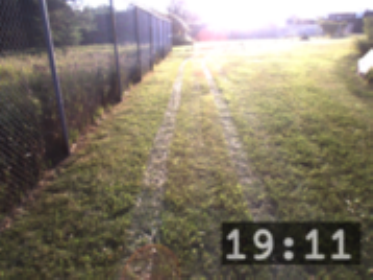}
  \end{subfigure}
  \begin{subfigure}[]{0.118\textwidth}
  		\centering
        \includegraphics[width=\textwidth]{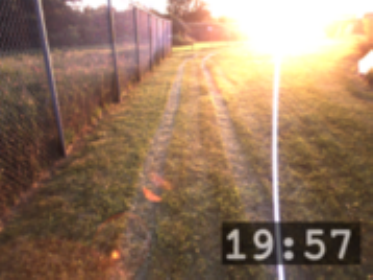}
  \end{subfigure}
%  \par\smallskip
  \begin{subfigure}[]{0.118\textwidth}
  		\centering
        \includegraphics[width=\textwidth]{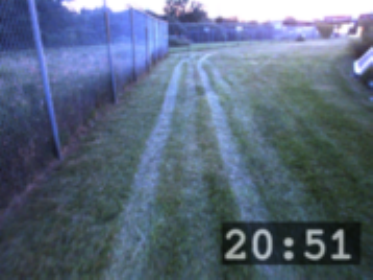}
  \end{subfigure}
  \begin{subfigure}[]{0.118\textwidth}
  		\centering
        \includegraphics[width=\textwidth]{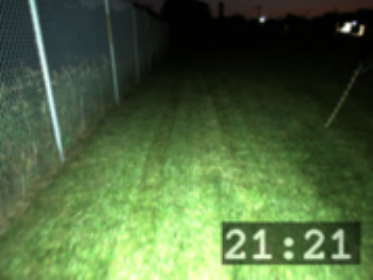}
  \end{subfigure}
  \begin{subfigure}[]{0.118\textwidth}
  		\centering
        \includegraphics[width=\textwidth]{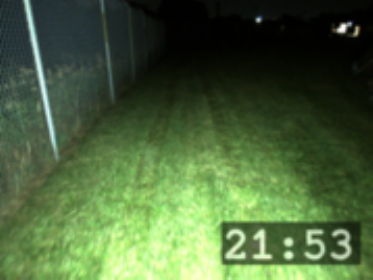}
  \end{subfigure}
  \begin{subfigure}[]{0.118\textwidth}
  		\centering
        \includegraphics[width=\textwidth]{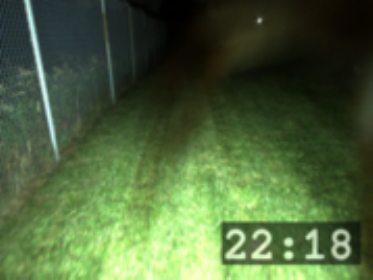}
  \end{subfigure}
  \caption{\small Images taken during closed-loop testing for the lighting-change experiment labeled with collection time (hh:mm).}
  \label{fig:lighting-change-images}
  \vspace{-0.25cm}
\end{figure*}

\begin{figure*}
  \centering
  \begin{subfigure}[]{0.118\textwidth}
  		\centering
        \includegraphics[width=\textwidth]{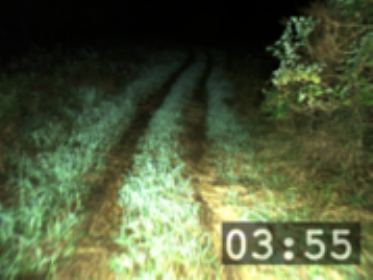}
  \end{subfigure}
  \begin{subfigure}[]{0.118\textwidth}
  		\centering
        \includegraphics[width=\textwidth]{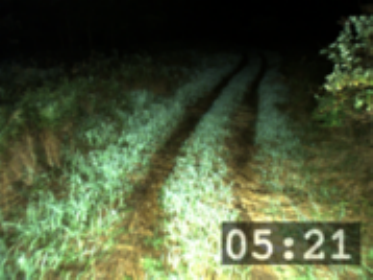}
  \end{subfigure}
  \begin{subfigure}[]{0.118\textwidth}
  		\centering
        \includegraphics[width=\textwidth]{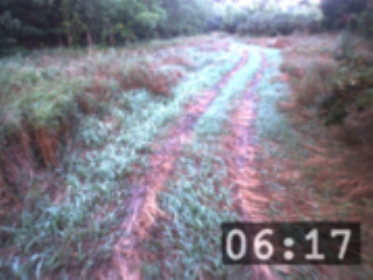}
  \end{subfigure}
  \begin{subfigure}[]{0.118\textwidth}
  		\centering
        \includegraphics[width=\textwidth]{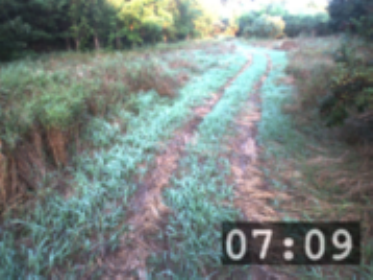}
  \end{subfigure}
  \begin{subfigure}[]{0.118\textwidth}
  		\centering
        \includegraphics[width=\textwidth]{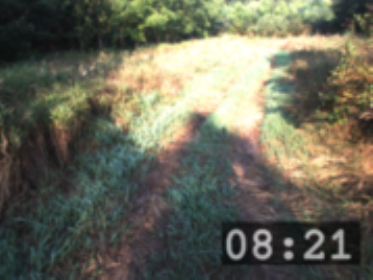}
  \end{subfigure}
  \begin{subfigure}[]{0.118\textwidth}
  		\centering
        \includegraphics[width=\textwidth]{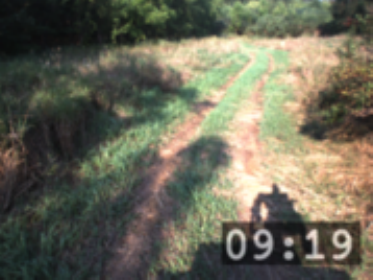}
  \end{subfigure}
  \begin{subfigure}[]{0.118\textwidth}
  		\centering
        \includegraphics[width=\textwidth]{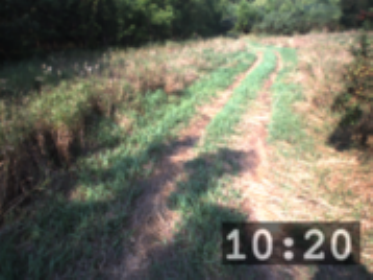}
  \end{subfigure}
  \begin{subfigure}[]{0.118\textwidth}
  		\centering
        \includegraphics[width=\textwidth]{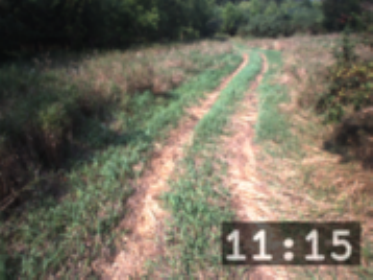}
  \end{subfigure}
  \par\smallskip
  \begin{subfigure}[]{0.118\textwidth}
  		\centering
        \includegraphics[width=\textwidth]{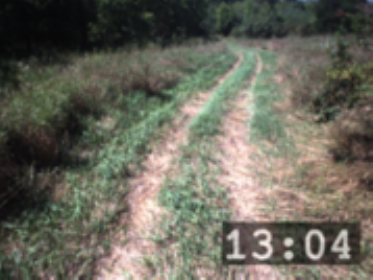}
  \end{subfigure}
  \begin{subfigure}[]{0.118\textwidth}
  		\centering
        \includegraphics[width=\textwidth]{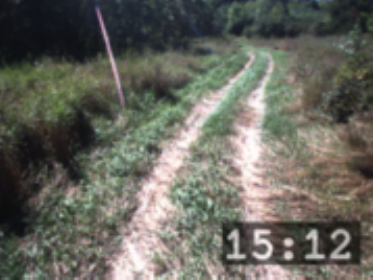}
  \end{subfigure}
  \begin{subfigure}[]{0.118\textwidth}
  		\centering
        \includegraphics[width=\textwidth]{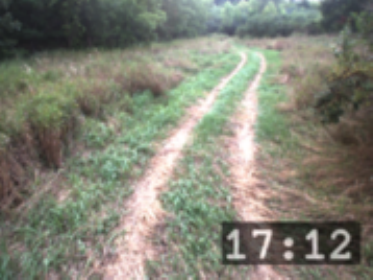}
  \end{subfigure}
  \begin{subfigure}[]{0.118\textwidth}
  		\centering
        \includegraphics[width=\textwidth]{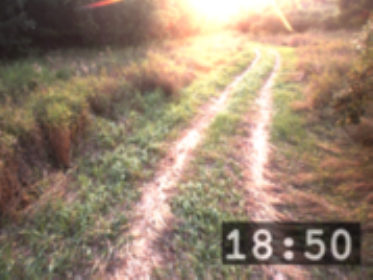}
  \end{subfigure}
  \begin{subfigure}[]{0.118\textwidth}
  		\centering
        \includegraphics[width=\textwidth]{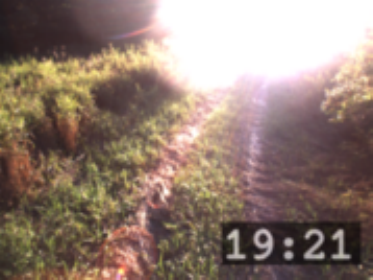}
  \end{subfigure}
  \begin{subfigure}[]{0.118\textwidth}
  		\centering
        \includegraphics[width=\textwidth]{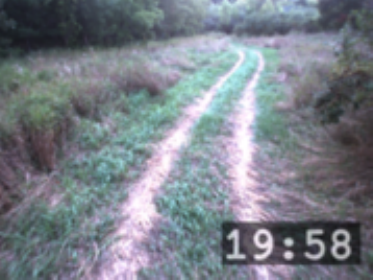}
  \end{subfigure}
  \begin{subfigure}[]{0.118\textwidth}
  		\centering
        \includegraphics[width=\textwidth]{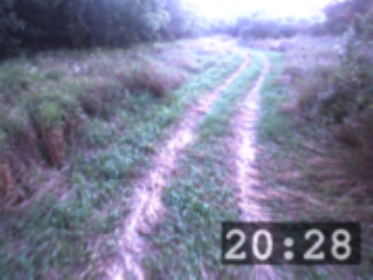}
  \end{subfigure}
  \begin{subfigure}[]{0.118\textwidth}
  		\centering
        \includegraphics[width=\textwidth]{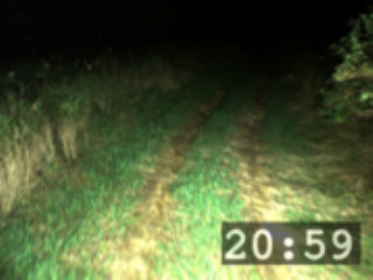}
  \end{subfigure}
  \caption{\small Images taken during closed-loop testing for the first generalization experiment labeled with collection time (hh:mm).}
  \label{fig:generalization-images}
  \vspace{-0.25cm}
\end{figure*}

\begin{figure*}
  \centering
  \begin{subfigure}[]{0.118\textwidth}
  		\centering
        \includegraphics[width=\textwidth]{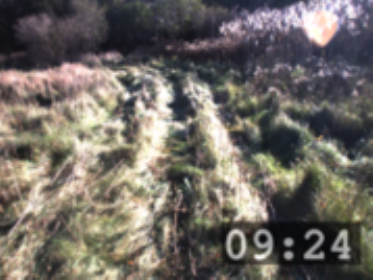}
  \end{subfigure}
  \begin{subfigure}[]{0.118\textwidth}
  		\centering
        \includegraphics[width=\textwidth]{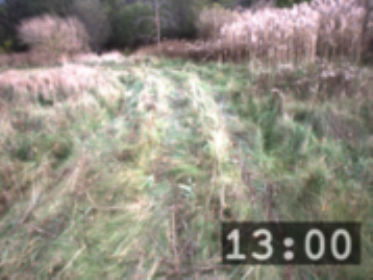}
  \end{subfigure}
  \begin{subfigure}[]{0.118\textwidth}
  		\centering
        \includegraphics[width=\textwidth]{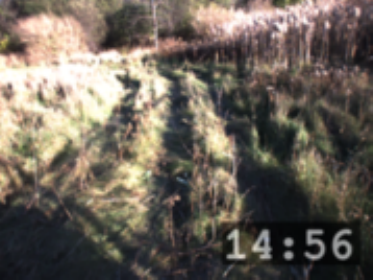}
  \end{subfigure}
  \begin{subfigure}[]{0.118\textwidth}
  		\centering
        \includegraphics[width=\textwidth]{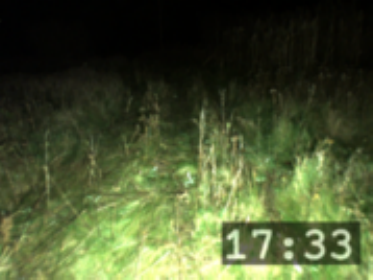}
  \end{subfigure}
  \begin{subfigure}[]{0.118\textwidth}
  		\centering
        \includegraphics[width=\textwidth]{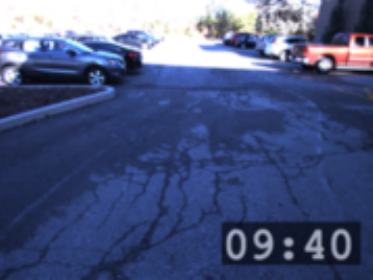}
  \end{subfigure}
  \begin{subfigure}[]{0.118\textwidth}
  		\centering
        \includegraphics[width=\textwidth]{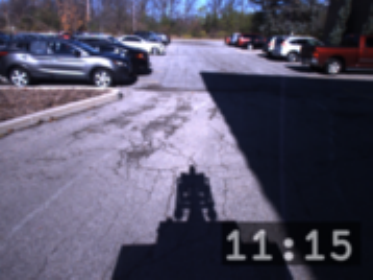}
  \end{subfigure}
  \begin{subfigure}[]{0.118\textwidth}
  		\centering
        \includegraphics[width=\textwidth]{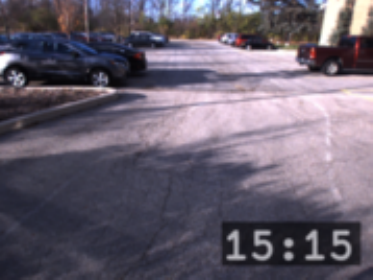}
  \end{subfigure}
  \begin{subfigure}[]{0.118\textwidth}
  		\centering
        \includegraphics[width=\textwidth]{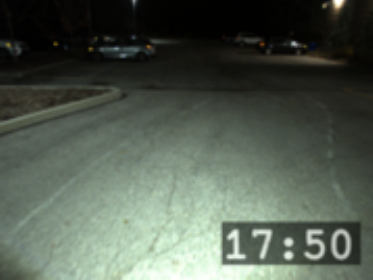}
  \end{subfigure}
  \caption{\small Images taken during closed-loop testing for the second (left four) and third (right four) generalization experiments labeled with collection time (hh:mm).}
  \label{fig:generalization2-images}
  \vspace{-0.5cm}
\end{figure*}

\setlength{\tabcolsep}{0.189em}
\begin{table*}[h]
{\renewcommand{\arraystretch}{1.1}
\begin{tabular}{cc|cccccccccccccccccccccccccccc|}
\cline{3-30}
 &                                                 & \multicolumn{28}{c|}{Repeat}                                                                                                                                                                                                                                                                                                                                                                                                                                                                                                                                                                                                                                                                                                                                                                                                                                                                                                                                                                                                                                                                                                                                                                                                                                                                                                                                                                                                                                                                                                                \\ \cline{3-30} 
                                                & \cellcolor[HTML]{FFFFFF}{\color[HTML]{333333} } & \multicolumn{4}{c|}{31/01}                                                                                                                                                                                                 & \multicolumn{4}{c|}{02/02}                                                                                                                                                                                                 & \multicolumn{4}{c|}{10/02}                                                                                                                                                                                                 & \multicolumn{4}{c|}{14/02}                                                                                                                                                                                      & \multicolumn{4}{c|}{27/02}                                                                                                                                                                                        & \multicolumn{4}{c|}{10/04}                                                                                                                                                                                        & \multicolumn{4}{c|}{03/05}                                                                                                                                                                 \\ \cline{3-30} 
\multicolumn{1}{l}{}                            & \multicolumn{1}{l|}{}                           & \multicolumn{1}{l|}{SP}                         & \multicolumn{1}{l|}{D2}                          & \multicolumn{1}{l|}{R2D2}                                 & \multicolumn{1}{l|}{Our}                                  & \multicolumn{1}{l|}{SP}                         & \multicolumn{1}{l|}{D2}                          & \multicolumn{1}{l|}{R2D2}                                 & \multicolumn{1}{l|}{Our}                                  & \multicolumn{1}{l|}{SP}                         & \multicolumn{1}{l|}{D2}                          & \multicolumn{1}{l|}{R2D2}                                 & \multicolumn{1}{l|}{Our}                                  & \multicolumn{1}{l|}{SP}                         & \multicolumn{1}{l|}{D2}                         & \multicolumn{1}{l|}{R2D2}                       & \multicolumn{1}{l|}{Our}                                  & \multicolumn{1}{l|}{SP}                         & \multicolumn{1}{l|}{D2}                          & \multicolumn{1}{l|}{R2D2}                        & \multicolumn{1}{l|}{Our}                                  & \multicolumn{1}{l|}{SP}                         & \multicolumn{1}{l|}{D2}                          & \multicolumn{1}{l|}{R2D2}                        & \multicolumn{1}{l|}{Our}                                  & \multicolumn{1}{l|}{SP}                         & \multicolumn{1}{l|}{D2}                         & \multicolumn{1}{l|}{R2D2}                       & \multicolumn{1}{l|}{Our}             \\ \hline

\multicolumn{1}{|c}{} \parbox[t]{2mm}{\multirow{7}{*}{\rotatebox[origin=c]{90}{Teach}}}                        & \multicolumn{1}{|c|}{31/01}                                            & \multicolumn{1}{c|}{\cellcolor[HTML]{C9E2E4}}   & \multicolumn{1}{c|}{\cellcolor[HTML]{C9E2E4}}    & \multicolumn{1}{c|}{\cellcolor[HTML]{C9E2E4}}             & \multicolumn{1}{c|}{\cellcolor[HTML]{C9E2E4}}             & \multicolumn{1}{c|}{\cellcolor[HTML]{C9E2E4}70} & \multicolumn{1}{c|}{\cellcolor[HTML]{C9E2E4}272} & \multicolumn{1}{c|}{\cellcolor[HTML]{C9E2E4}\textbf{613}} & \multicolumn{1}{c|}{\cellcolor[HTML]{C9E2E4}311}          & \multicolumn{1}{c|}{\cellcolor[HTML]{C9E2E4}35} & \multicolumn{1}{c|}{\cellcolor[HTML]{C9E2E4}76}  & \multicolumn{1}{c|}{\cellcolor[HTML]{C9E2E4}\textbf{284}} & \multicolumn{1}{c|}{\cellcolor[HTML]{C9E2E4}230}          & \multicolumn{1}{c|}{\cellcolor[HTML]{C9E2E4}4}  & \multicolumn{1}{c|}{\cellcolor[HTML]{C9E2E4}2}  & \multicolumn{1}{c|}{\cellcolor[HTML]{C9E2E4}4}  & \multicolumn{1}{c|}{\cellcolor[HTML]{C9E2E4}\textbf{112}} & \multicolumn{1}{c|}{\cellcolor[HTML]{C9E2E4}4}  & \multicolumn{1}{c|}{\cellcolor[HTML]{C9E2E4}46}  & \multicolumn{1}{c|}{\cellcolor[HTML]{C9E2E4}17}  & \multicolumn{1}{c|}{\cellcolor[HTML]{C9E2E4}\textbf{201}} & \multicolumn{1}{c|}{\cellcolor[HTML]{C9E2E4}4}  & \multicolumn{1}{c|}{\cellcolor[HTML]{C9E2E4}37}  & \multicolumn{1}{c|}{\cellcolor[HTML]{C9E2E4}7}   & \multicolumn{1}{c|}{\cellcolor[HTML]{C9E2E4}\textbf{186}} & \multicolumn{1}{c|}{\cellcolor[HTML]{C9E2E4}4}  & \multicolumn{1}{c|}{\cellcolor[HTML]{C9E2E4}39} & \multicolumn{1}{c|}{\cellcolor[HTML]{C9E2E4}9}  & \cellcolor[HTML]{C9E2E4}\textbf{154} \\ \cline{2-2}

\multicolumn{1}{|c}{}                          & \multicolumn{1}{|c|}{02/02}                                           & \multicolumn{1}{c|}{\cellcolor[HTML]{FFFFFF}69} & \multicolumn{1}{c|}{\cellcolor[HTML]{FFFFFF}263} & \multicolumn{1}{c|}{\cellcolor[HTML]{FFFFFF}\textbf{606}} & \multicolumn{1}{c|}{\cellcolor[HTML]{FFFFFF}310}          & \multicolumn{1}{c|}{\cellcolor[HTML]{FFFFFF}}   & \multicolumn{1}{c|}{\cellcolor[HTML]{FFFFFF}}    & \multicolumn{1}{c|}{\cellcolor[HTML]{FFFFFF}}             & \multicolumn{1}{c|}{\cellcolor[HTML]{FFFFFF}}             & \multicolumn{1}{c|}{\cellcolor[HTML]{FFFFFF}77} & \multicolumn{1}{c|}{\cellcolor[HTML]{FFFFFF}154} & \multicolumn{1}{c|}{\cellcolor[HTML]{FFFFFF}\textbf{319}} & \multicolumn{1}{c|}{\cellcolor[HTML]{FFFFFF}248}          & \multicolumn{1}{c|}{\cellcolor[HTML]{FFFFFF}16} & \multicolumn{1}{c|}{\cellcolor[HTML]{FFFFFF}17} & \multicolumn{1}{c|}{\cellcolor[HTML]{FFFFFF}10} & \multicolumn{1}{c|}{\cellcolor[HTML]{FFFFFF}\textbf{124}} & \multicolumn{1}{c|}{\cellcolor[HTML]{FFFFFF}11} & \multicolumn{1}{c|}{\cellcolor[HTML]{FFFFFF}47}  & \multicolumn{1}{c|}{\cellcolor[HTML]{FFFFFF}21}  & \multicolumn{1}{c|}{\cellcolor[HTML]{FFFFFF}\textbf{223}} & \multicolumn{1}{c|}{\cellcolor[HTML]{FFFFFF}6}  & \multicolumn{1}{c|}{\cellcolor[HTML]{FFFFFF}36}  & \multicolumn{1}{c|}{\cellcolor[HTML]{FFFFFF}5}   & \multicolumn{1}{c|}{\cellcolor[HTML]{FFFFFF}\textbf{206}} & \multicolumn{1}{c|}{\cellcolor[HTML]{FFFFFF}8}  & \multicolumn{1}{c|}{\cellcolor[HTML]{FFFFFF}34} & \multicolumn{1}{c|}{\cellcolor[HTML]{FFFFFF}5}  & \cellcolor[HTML]{FFFFFF}\textbf{166} \\ \cline{2-2}

\multicolumn{1}{|c}{}                          & \multicolumn{1}{|c|}{10/02}                                           & \multicolumn{1}{c|}{\cellcolor[HTML]{C9E2E4}31} & \multicolumn{1}{c|}{\cellcolor[HTML]{C9E2E4}76}  & \multicolumn{1}{c|}{\cellcolor[HTML]{C9E2E4}\textbf{283}} & \multicolumn{1}{c|}{\cellcolor[HTML]{C9E2E4}230}          & \multicolumn{1}{c|}{\cellcolor[HTML]{C9E2E4}79} & \multicolumn{1}{c|}{\cellcolor[HTML]{C9E2E4}155} & \multicolumn{1}{c|}{\cellcolor[HTML]{C9E2E4}\textbf{313}} & \multicolumn{1}{c|}{\cellcolor[HTML]{C9E2E4}248}          & \multicolumn{1}{c|}{\cellcolor[HTML]{C9E2E4}}   & \multicolumn{1}{c|}{\cellcolor[HTML]{C9E2E4}}    & \multicolumn{1}{c|}{\cellcolor[HTML]{C9E2E4}}             & \multicolumn{1}{c|}{\cellcolor[HTML]{C9E2E4}}             & \multicolumn{1}{c|}{\cellcolor[HTML]{C9E2E4}24} & \multicolumn{1}{c|}{\cellcolor[HTML]{C9E2E4}22} & \multicolumn{1}{c|}{\cellcolor[HTML]{C9E2E4}30} & \multicolumn{1}{c|}{\cellcolor[HTML]{C9E2E4}\textbf{117}} & \multicolumn{1}{c|}{\cellcolor[HTML]{C9E2E4}3}  & \multicolumn{1}{c|}{\cellcolor[HTML]{C9E2E4}10}  & \multicolumn{1}{c|}{\cellcolor[HTML]{C9E2E4}3}   & \multicolumn{1}{c|}{\cellcolor[HTML]{C9E2E4}\textbf{173}} & \multicolumn{1}{c|}{\cellcolor[HTML]{C9E2E4}3}  & \multicolumn{1}{c|}{\cellcolor[HTML]{C9E2E4}8}   & \multicolumn{1}{c|}{\cellcolor[HTML]{C9E2E4}2}   & \multicolumn{1}{c|}{\cellcolor[HTML]{C9E2E4}\textbf{161}} & \multicolumn{1}{c|}{\cellcolor[HTML]{C9E2E4}2}  & \multicolumn{1}{c|}{\cellcolor[HTML]{C9E2E4}8}  & \multicolumn{1}{c|}{\cellcolor[HTML]{C9E2E4}2}  & \cellcolor[HTML]{C9E2E4}\textbf{131} \\ \cline{2-2}

\multicolumn{1}{|c}{}                          & \multicolumn{1}{|c|}{14/02}                                           & \multicolumn{1}{c|}{\cellcolor[HTML]{FFFFFF}5}  & \multicolumn{1}{c|}{\cellcolor[HTML]{FFFFFF}2}   & \multicolumn{1}{c|}{\cellcolor[HTML]{FFFFFF}3}            & \multicolumn{1}{c|}{\cellcolor[HTML]{FFFFFF}\textbf{114}} & \multicolumn{1}{c|}{\cellcolor[HTML]{FFFFFF}19} & \multicolumn{1}{c|}{\cellcolor[HTML]{FFFFFF}31}  & \multicolumn{1}{c|}{\cellcolor[HTML]{FFFFFF}15}           & \multicolumn{1}{c|}{\cellcolor[HTML]{FFFFFF}\textbf{134}} & \multicolumn{1}{c|}{\cellcolor[HTML]{FFFFFF}30} & \multicolumn{1}{c|}{\cellcolor[HTML]{FFFFFF}31}  & \multicolumn{1}{c|}{\cellcolor[HTML]{FFFFFF}45}           & \multicolumn{1}{c|}{\cellcolor[HTML]{FFFFFF}\textbf{128}} & \multicolumn{1}{c|}{\cellcolor[HTML]{FFFFFF}}   & \multicolumn{1}{c|}{\cellcolor[HTML]{FFFFFF}}   & \multicolumn{1}{c|}{\cellcolor[HTML]{FFFFFF}}   & \multicolumn{1}{c|}{\cellcolor[HTML]{FFFFFF}\textbf{}}    & \multicolumn{1}{c|}{\cellcolor[HTML]{FFFFFF}2}  & \multicolumn{1}{c|}{\cellcolor[HTML]{FFFFFF}2}   & \multicolumn{1}{c|}{\cellcolor[HTML]{FFFFFF}3}   & \multicolumn{1}{c|}{\cellcolor[HTML]{FFFFFF}\textbf{121}} & \multicolumn{1}{c|}{\cellcolor[HTML]{FFFFFF}1}  & \multicolumn{1}{c|}{\cellcolor[HTML]{FFFFFF}0}   & \multicolumn{1}{c|}{\cellcolor[HTML]{FFFFFF}0}   & \multicolumn{1}{c|}{\cellcolor[HTML]{FFFFFF}\textbf{111}} & \multicolumn{1}{c|}{\cellcolor[HTML]{FFFFFF}1}  & \multicolumn{1}{c|}{\cellcolor[HTML]{FFFFFF}1}  & \multicolumn{1}{c|}{\cellcolor[HTML]{FFFFFF}0}  & \cellcolor[HTML]{FFFFFF}\textbf{102} \\ \cline{2-2}

\multicolumn{1}{|c}{}                          & \multicolumn{1}{|c|}{27/02}                                           & \multicolumn{1}{c|}{\cellcolor[HTML]{C9E2E4}4}  & \multicolumn{1}{c|}{\cellcolor[HTML]{C9E2E4}45}  & \multicolumn{1}{c|}{\cellcolor[HTML]{C9E2E4}15}           & \multicolumn{1}{c|}{\cellcolor[HTML]{C9E2E4}\textbf{201}} & \multicolumn{1}{c|}{\cellcolor[HTML]{C9E2E4}12} & \multicolumn{1}{c|}{\cellcolor[HTML]{C9E2E4}47}  & \multicolumn{1}{c|}{\cellcolor[HTML]{C9E2E4}22}           & \multicolumn{1}{c|}{\cellcolor[HTML]{C9E2E4}\textbf{223}} & \multicolumn{1}{c|}{\cellcolor[HTML]{C9E2E4}4}  & \multicolumn{1}{c|}{\cellcolor[HTML]{C9E2E4}13}  & \multicolumn{1}{c|}{\cellcolor[HTML]{C9E2E4}5}            & \multicolumn{1}{c|}{\cellcolor[HTML]{C9E2E4}\textbf{171}} & \multicolumn{1}{c|}{\cellcolor[HTML]{C9E2E4}2}  & \multicolumn{1}{c|}{\cellcolor[HTML]{C9E2E4}2}  & \multicolumn{1}{c|}{\cellcolor[HTML]{C9E2E4}2}  & \multicolumn{1}{c|}{\cellcolor[HTML]{C9E2E4}\textbf{114}} & \multicolumn{1}{c|}{\cellcolor[HTML]{C9E2E4}}   & \multicolumn{1}{c|}{\cellcolor[HTML]{C9E2E4}}    & \multicolumn{1}{c|}{\cellcolor[HTML]{C9E2E4}}    & \multicolumn{1}{c|}{\cellcolor[HTML]{C9E2E4}\textbf{}}    & \multicolumn{1}{c|}{\cellcolor[HTML]{C9E2E4}15} & \multicolumn{1}{c|}{\cellcolor[HTML]{C9E2E4}136} & \multicolumn{1}{c|}{\cellcolor[HTML]{C9E2E4}104} & \multicolumn{1}{c|}{\cellcolor[HTML]{C9E2E4}\textbf{257}} & \multicolumn{1}{c|}{\cellcolor[HTML]{C9E2E4}13} & \multicolumn{1}{c|}{\cellcolor[HTML]{C9E2E4}75} & \multicolumn{1}{c|}{\cellcolor[HTML]{C9E2E4}25} & \cellcolor[HTML]{C9E2E4}\textbf{196} \\ \cline{2-2}

\multicolumn{1}{|c}{}                          & \multicolumn{1}{|c|}{10/04}                                           & \multicolumn{1}{c|}{\cellcolor[HTML]{FFFFFF}3}  & \multicolumn{1}{c|}{\cellcolor[HTML]{FFFFFF}34}  & \multicolumn{1}{c|}{\cellcolor[HTML]{FFFFFF}7}            & \multicolumn{1}{c|}{\cellcolor[HTML]{FFFFFF}\textbf{183}} & \multicolumn{1}{c|}{\cellcolor[HTML]{FFFFFF}7}  & \multicolumn{1}{c|}{\cellcolor[HTML]{FFFFFF}31}  & \multicolumn{1}{c|}{\cellcolor[HTML]{FFFFFF}4}            & \multicolumn{1}{c|}{\cellcolor[HTML]{FFFFFF}\textbf{206}} & \multicolumn{1}{c|}{\cellcolor[HTML]{FFFFFF}3}  & \multicolumn{1}{c|}{\cellcolor[HTML]{FFFFFF}7}   & \multicolumn{1}{c|}{\cellcolor[HTML]{FFFFFF}1}            & \multicolumn{1}{c|}{\cellcolor[HTML]{FFFFFF}\textbf{158}} & \multicolumn{1}{c|}{\cellcolor[HTML]{FFFFFF}1}  & \multicolumn{1}{c|}{\cellcolor[HTML]{FFFFFF}0}  & \multicolumn{1}{c|}{\cellcolor[HTML]{FFFFFF}0}  & \multicolumn{1}{c|}{\cellcolor[HTML]{FFFFFF}\textbf{106}} & \multicolumn{1}{c|}{\cellcolor[HTML]{FFFFFF}13} & \multicolumn{1}{c|}{\cellcolor[HTML]{FFFFFF}136} & \multicolumn{1}{c|}{\cellcolor[HTML]{FFFFFF}119} & \multicolumn{1}{c|}{\cellcolor[HTML]{FFFFFF}\textbf{257}} & \multicolumn{1}{c|}{\cellcolor[HTML]{FFFFFF}}   & \multicolumn{1}{c|}{\cellcolor[HTML]{FFFFFF}}    & \multicolumn{1}{c|}{\cellcolor[HTML]{FFFFFF}}    & \multicolumn{1}{c|}{\cellcolor[HTML]{FFFFFF}\textbf{}}    & \multicolumn{1}{c|}{\cellcolor[HTML]{FFFFFF}15} & \multicolumn{1}{c|}{\cellcolor[HTML]{FFFFFF}56} & \multicolumn{1}{c|}{\cellcolor[HTML]{FFFFFF}33} & \cellcolor[HTML]{FFFFFF}\textbf{208} \\ \cline{2-2}

\multicolumn{1}{|c}{\multirow{-7}{*}{}}        & \multicolumn{1}{|c|}{03/05}                                           & \multicolumn{1}{c|}{\cellcolor[HTML]{C9E2E4}4}  & \multicolumn{1}{c|}{\cellcolor[HTML]{C9E2E4}36}  & \multicolumn{1}{c|}{\cellcolor[HTML]{C9E2E4}9}            & \multicolumn{1}{c|}{\cellcolor[HTML]{C9E2E4}\textbf{153}} & \multicolumn{1}{c|}{\cellcolor[HTML]{C9E2E4}8}  & \multicolumn{1}{c|}{\cellcolor[HTML]{C9E2E4}38}  & \multicolumn{1}{c|}{\cellcolor[HTML]{C9E2E4}4}            & \multicolumn{1}{c|}{\cellcolor[HTML]{C9E2E4}\textbf{166}} & \multicolumn{1}{c|}{\cellcolor[HTML]{C9E2E4}3}  & \multicolumn{1}{c|}{\cellcolor[HTML]{C9E2E4}10}  & \multicolumn{1}{c|}{\cellcolor[HTML]{C9E2E4}2}            & \multicolumn{1}{c|}{\cellcolor[HTML]{C9E2E4}\textbf{131}} & \multicolumn{1}{c|}{\cellcolor[HTML]{C9E2E4}1}  & \multicolumn{1}{c|}{\cellcolor[HTML]{C9E2E4}2}  & \multicolumn{1}{c|}{\cellcolor[HTML]{C9E2E4}0}  & \multicolumn{1}{c|}{\cellcolor[HTML]{C9E2E4}\textbf{98}}  & \multicolumn{1}{c|}{\cellcolor[HTML]{C9E2E4}14} & \multicolumn{1}{c|}{\cellcolor[HTML]{C9E2E4}74}  & \multicolumn{1}{c|}{\cellcolor[HTML]{C9E2E4}25}  & \multicolumn{1}{c|}{\cellcolor[HTML]{C9E2E4}\textbf{197}} & \multicolumn{1}{c|}{\cellcolor[HTML]{C9E2E4}15} & \multicolumn{1}{c|}{\cellcolor[HTML]{C9E2E4}55}  & \multicolumn{1}{c|}{\cellcolor[HTML]{C9E2E4}33}  & \multicolumn{1}{c|}{\cellcolor[HTML]{C9E2E4}\textbf{207}} & \multicolumn{1}{c|}{\cellcolor[HTML]{C9E2E4}}   & \multicolumn{1}{c|}{\cellcolor[HTML]{C9E2E4}}   & \multicolumn{1}{c|}{\cellcolor[HTML]{C9E2E4}}   & \cellcolor[HTML]{C9E2E4}             \\ \hline
\end{tabular}}
\caption{\small Comparison of the median number of inliers for SuperPoint, D2-Net, R2D2, and our method for localization of the runs shown in Figure \ref{fig:offline_images}. The rows list the condition used to teach by date (dd/mm), while the columns correspond to the repeats. As localization becomes more challenging, our method keeps a high number of inliers, while the other methods deteriorate. }
\label{tab:comparison}
\vspace{-0.4cm}
\end{table*}

\setlength{\tabcolsep}{0.58em}
\begin{table}[]
\begin{tabular}{c@{\hspace{-0.3 mm}} r  c @{\hspace{-0.3 mm}} c r }
\hline
Path             & \multicolumn{1}{c}{Total dist.} & Runs w. fail        & Num. fail & \multicolumn{1}{c}{Dist. VO} \\ \hline
Lighting Change  & 10.9 km                         & 22:18               & 3         & 0.4 m                        \\
Generalization 1 & 20.6 km                         & 03:55, 05:21, 20:59 & 364       & 29.2 m                       \\
Generalization 2 & 2.2 km                          & 14:56, 17:33        & 11        & 2.0 m                        \\
Generalization 3 & 1.8 km                          & -                   & 0         & 0.0 m                        \\ \hline
\end{tabular}
\caption{\small An overview of localization failures (less than 6 match inliers) in different experiments. The last column shows the total distance driven on VO during localization failures.}
\label{tab:loc_failure}
\vspace{-0.5cm}
\end{table}

\begin{figure*}[t]
  \centering
  \begin{subfigure}[]{1.0\textwidth}
  		\centering
        \includegraphics[width=\textwidth,trim={0 0.35cm 0 0},clip]{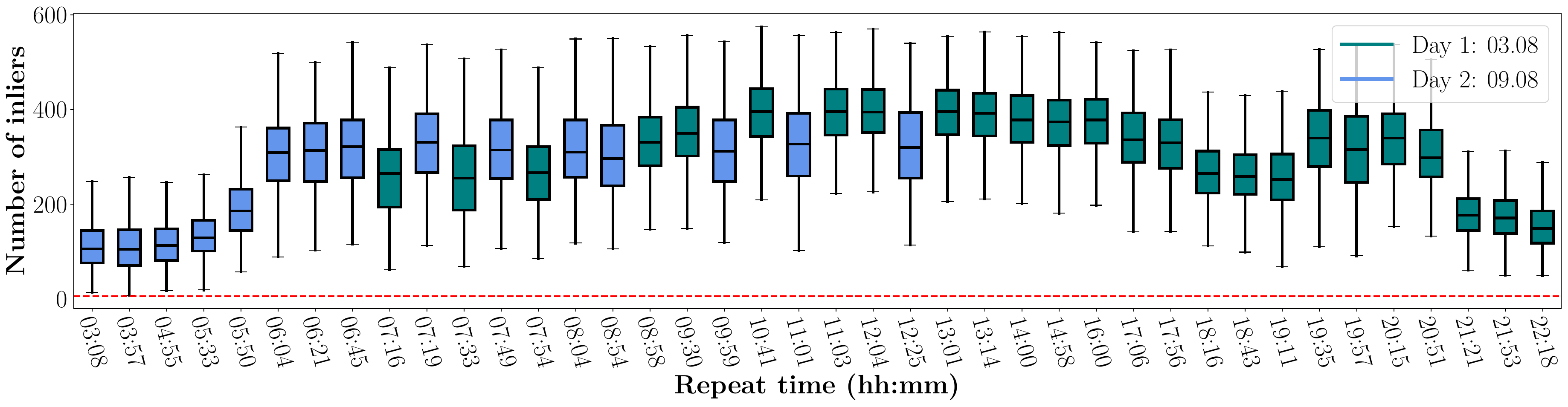}
  \end{subfigure}
  \vspace{-2 mm}
  \caption{\small Plot of inliers for the lighting-change experiment. The red line shows the minimum number of inliers (6) required.}
  \label{fig:inlier_plots_light}
  \vspace{-0.4cm}
\end{figure*}

\section{Experiments}

\subsection{Data}

The training data were previously collected using a Clearpath Grizzly robot with a Bumblebee XB3 camera, see Figure \ref{fig:overview}. By using multi-experience \ac{VTR} \cite{Paton2016}, the robot repeats paths accurately across large lighting and seasonal change with the help of intermediate bridging experiences and \ac{SURF} \cite{Bay2006}. We can use the resulting data as ground truth for supervised learning. In \ac{VTR}, stereo image keyframes are stored as vertices in a spatio-temporal pose graph. Edges contain the relative pose between temporally adjacent vertices and between a repeat vertex and the teach vertex to which it has been localized. We sample image pairs and poses from the pose graph. 

We use data from two different paths for training\footnote{Data at http://asrl.utias.utoronto.ca/datasets/2020-vtr-dataset}. The In-The-Dark dataset contains 39 runs of a path collected at our campus in summer 2016 along a road and on grass. The path was repeated once per hour for 30 hours systematically capturing lighting change. The Multiseason dataset contains 136 runs of a path in an area with more vegetation and undulating terrain. It was repeated from January until May 2017 capturing varying seasons and weather. Overhead path views can be seen in Figure \ref{fig:paths}. We generate two separate datasets from these two paths that each have 100,000 training samples and 20,000 validation samples. 

\subsection{Training and Inference}

We train the network by giving it pairs of images and the ground truth relative pose from the training dataset. Large outliers are removed based on keypoint error using the ground truth pose during training and with \ac{RANSAC} during inference. The network is trained using the Adam optimizer \cite{Kingma2014} with a learning rate of $10^{-5}$ and other parameters set to default values. We determine the number of training epochs with early stopping based on the validation loss. The network is trained on an NVIDIA Tesla V100 DGXS GPU. Feature extraction on this server using PyTorch takes on average 14.8 ms, while it takes 7.3 ms using C++ in the VT\&R system running on a ThinkPad P52 laptop with an Intel \textregistered \space Core\textsuperscript{\texttrademark} i7-8850H CPU, 32 GB of RAM, and an NVIDIA Quadro P2000 4 GB GPU. The code is made available online \footnote{Code at https://github.com/utiasASRL/deep\_learned\_visual\_features}. 

\subsection{Visual Teach and Repeat}
\label{sec:vtr}

In order to test the performance of the learned features, we add them to the \ac{VTR} system. While a user manually drives the robot to teach a new path, \ac{VTR} creates a spatio-temporal pose graph that serves as the map. The path is repeated by alternating between VO and localizing keyframes. VT\&R does not compute global poses, but only finds the relative offset to a keyframe in the map. In multi-experience localization, poses and detected features for each repeat is stored in the pose graph. More detail on \ac{VTR} can be found in \cite{Paton2018}, and the \ac{VTR} code base is available online\footnote{VT\&R code at http://utiasasrl.github.io/vtr3}. We insert the learned feature detector and descriptors without making substantial changes to \ac{VTR}. Instead, we use the existing sparse descriptor matcher and do not add the keypoint scores used in training, but plan to include dense descriptor matching and keypoint scores in future work. Note that in the experiments, we match live images directly to the map without using intermediate experiences. Since VT\&R still relies on SURF for VO, the learned features introduce additional computation and we had to reduce the maximum speed of the robot to 0.8 m/s in order to run the experiment.

We complete three experiments. We start by comparing to other state-of-the-art methods with code available online, specifically SuperPoint \cite{detone2018a}, D2-Net \cite{dusmanu2019}, and R2D2 \cite{revaud2019}. This comparison is done offline in PyTorch (not in VT\&R) for localization only on held-out repeats from the Multiseason dataset, see Figure \ref{fig:offline_images}. Our method uses a network trained on the Multiseason and In-The-Dark datasets, while we use weights provided with the other methods.

The last two experiments are run in closed loop on the robot. First we train a network on the In-The-Dark dataset and teach a new path by physically driving the robot, rather than using held-out data from the training dataset. The new path is similar to the one form the training data (see Figure \ref{fig:live_paths}), albeit five years later. We will refer to this as the `lighting-change experiment'. We taught the approximately 265 m path at 12:14 on August 2nd and repeated it on August 3rd and 9th covering lighting conditions from 3 a.m. til 10.30 p.m., see Figure \ref{fig:lighting-change-images}. For this experiment, the network had a higher number of features for each layer (first layer of size 32 instead of 16). We later found this was unnecessary and all other experiments use the architecture from Figure \ref{fig:network}. 

In the final closed-loop experiments, which we refer to as the `generalization experiment', we test the learned features' ability to generalize to new areas. We train a network using both the Multiseason and In-The-Dark datasets and teach three new paths. The first is approximately 760 m and, as shown in Figure \ref{fig:live_paths}, includes two new areas that are not observed by training data. The larger of the two new sections is driven in an area with more vegetation, trees, and taller grass than seen in the most relevant Multiseason dataset. Moreover, the experiment is conducted in August, while the Multiseason dataset only contains data until May. This path was taught on August 14th and repeated on August 15th, 16th, and 20th covering lighting change from 4 a.m. til 9 p.m., see Figure \ref{fig:generalization-images}. Additionally, we include a second path taught in a field at 09:52 on November 11th (275 m) and a third path taught in a parking lot at 11:34 on November 11th (220 m). Neither path has any overlap with the training data, see Figure \ref{fig:live_paths}. The paths are repeated from morning until dark on November 11th and 12th, see Figure \ref{fig:generalization2-images}.        

\begin{figure*}[t]
  \centering
  \begin{subfigure}[]{1.0\textwidth}
		\centering        
        \includegraphics[width=\textwidth,trim={0 0.35cm 0 0},clip]{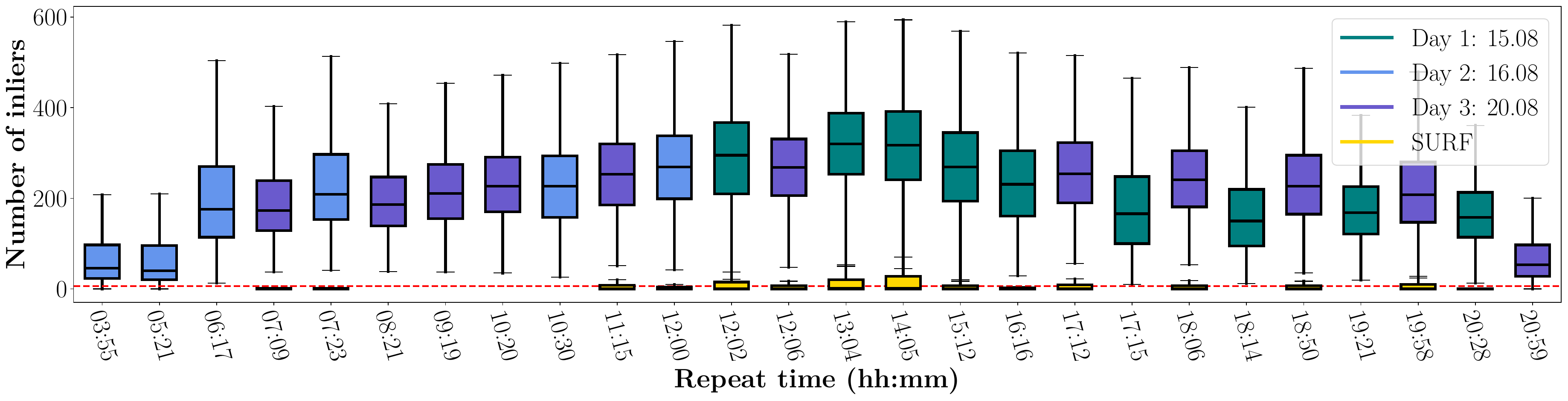}
        \vspace{-5.5 mm}
        \caption{\small Inliers for the first generalization experiment compared to SURF. Missing data are runs where SURF failed to localize from the start.}
  \end{subfigure}
  \begin{subfigure}[]{1.0\textwidth}
  		\centering
        \includegraphics[width=\textwidth,trim={0 0.35cm 0 0},clip]{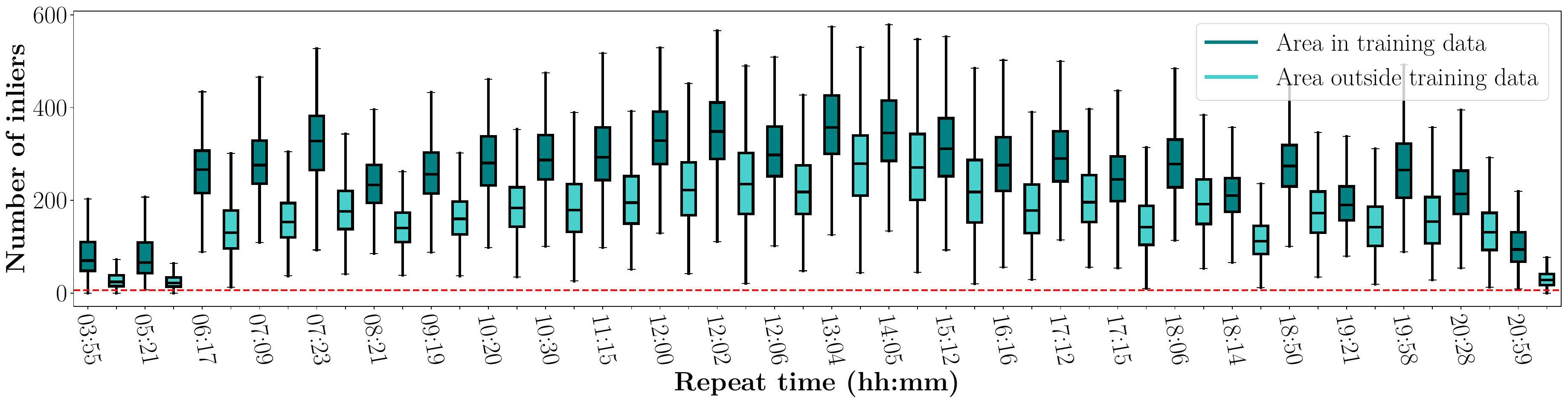}
        \vspace{-6 mm}
        \caption{\small Compared to the off-road area seen in training, the number of inliers in the new off-road area is lower but high enough for localization.}
  \end{subfigure}
  \begin{subfigure}[]{1.0\textwidth}
		\centering        
        \includegraphics[width=\textwidth,trim={0 0.35cm 0 0},clip]{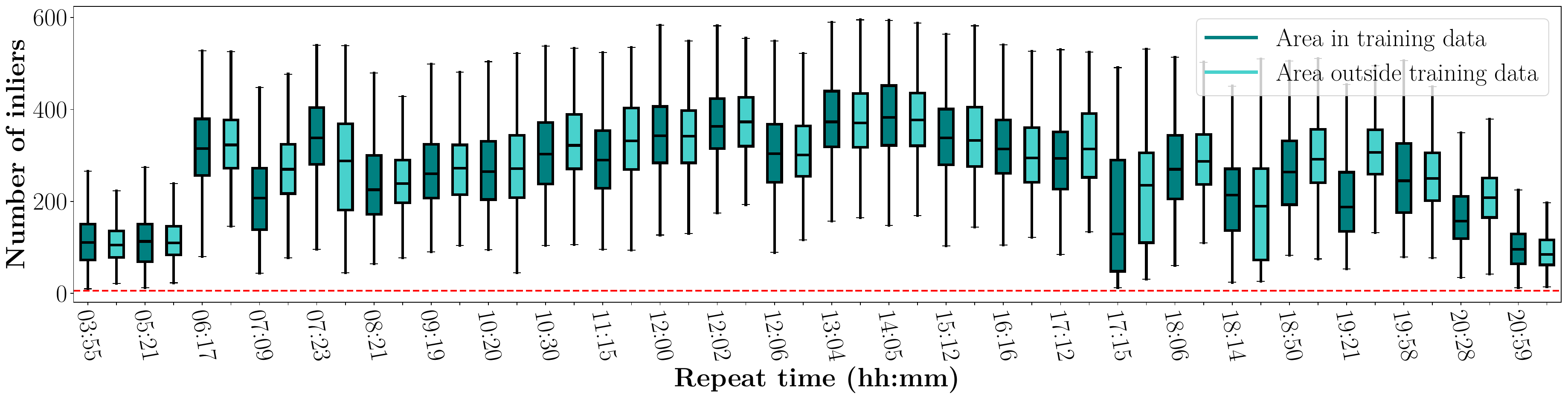}
        \vspace{-6 mm}
        \caption{\small Compared to the on-road area seen in training, the number of inliers in the new on-road area is similar or slightly higher.}
  \end{subfigure}
  \begin{subfigure}[]{0.49\textwidth}
		\centering        
        \includegraphics[width=\textwidth,trim={0 0.37cm 0 0},clip]{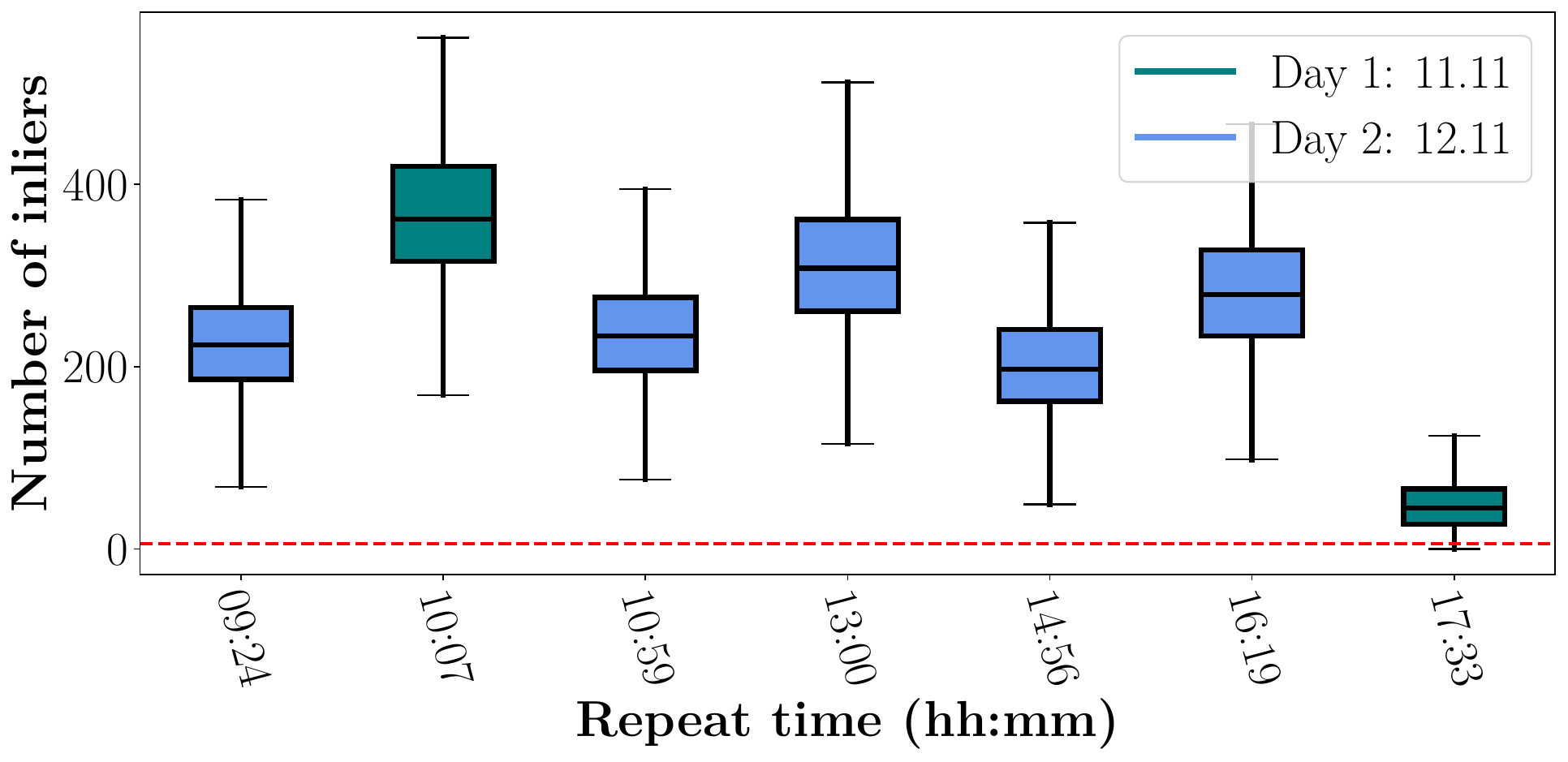}
        \vspace{-5.5 mm}
        \caption{\small Inliers for the second generalization experiment.}
  \end{subfigure}
  \begin{subfigure}[]{0.49\textwidth}
		\centering        
        \includegraphics[width=\textwidth,trim={0 0.37cm 0 0},clip]{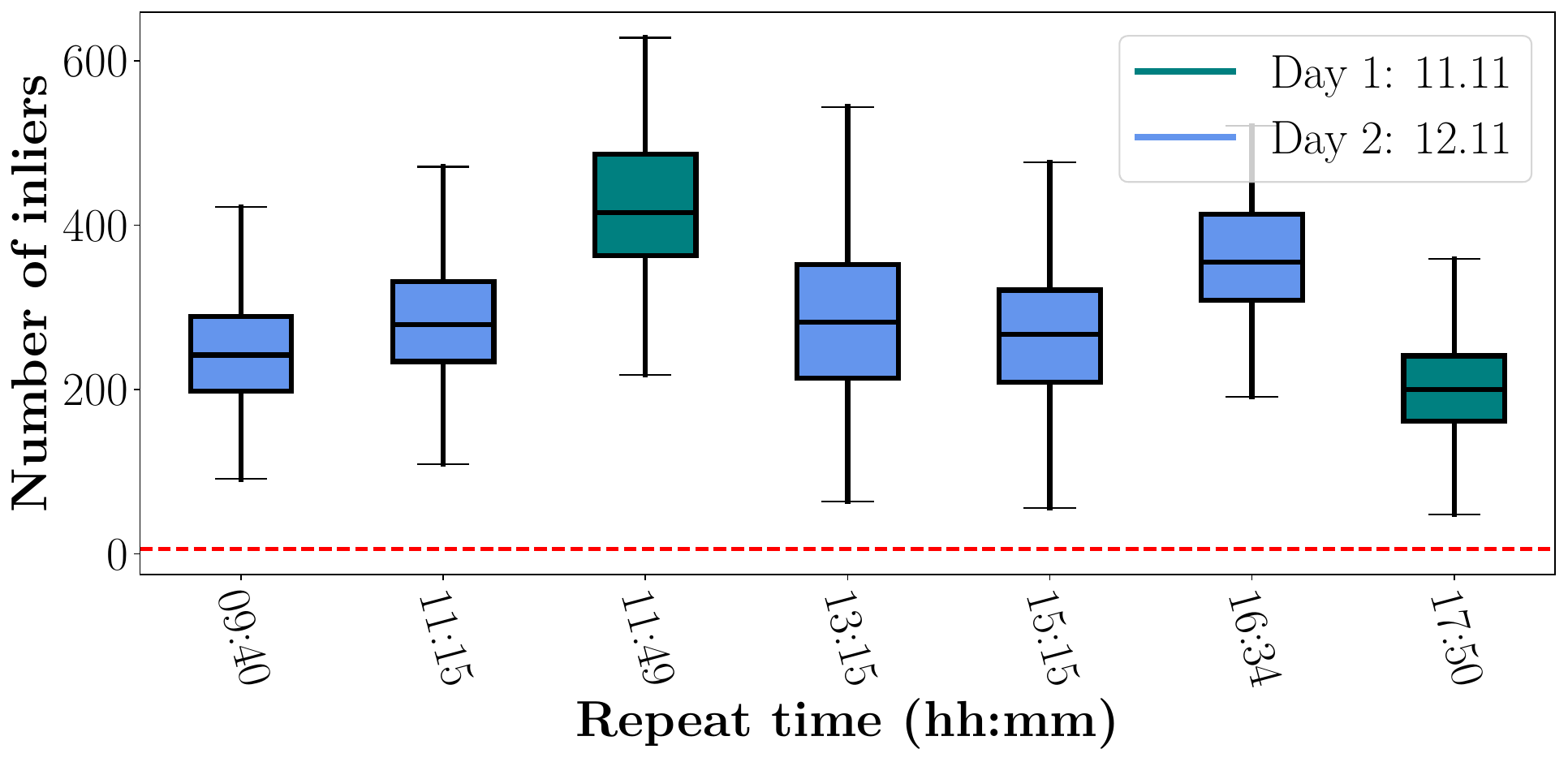}
        \vspace{-5.5 mm}
        \caption{\small Inliers for the third generalization experiment.}
  \end{subfigure}
  \vspace{-1 mm}
  \caption{\small Inliers for the generalization experiments. The red line shows the minimum number of inliers (6) required.}
  \label{fig:inlier_plots}
  \vspace{-0.6cm}
\end{figure*}

\section{Results}

\subsection{Offline: Learned Features Comparison }

We compared our method to state-of-the-art learned features on localization only in PyTorch. Keypoints, descriptors, and scores were extracted using the code provided by the authors, and the same sparse nearest-neighbour feature matching was used for all methods for fairness. For D2-Net and R2D2, we used multi-scale detection for best results. All conditions from Figure \ref{fig:offline_images} were localized against each other. We did this test as a sanity check as our method trained on relevant data should outperform the others \footnote{Since our data is different from more commonly used urban datasets, retraining the other methods with the same data would be more fair. This limits the usefulness of our comparison.}.

The median number of inliers as determined by RANSAC are listed in Table \ref{tab:comparison}. R2D2 gets the best result localizing the first three conditions to each other. Our method, however, is the only one with a high number of inliers across all experiments, while the others deteriorate as the seasonal change increases. Table \ref{tab:comparison} shows that the run from 14/02 with snow and high brightness is the most difficult. Since our target application runs on a laptop in real-time, memory consumption and speed also matter. R2D2 (1531 MB) and our method (2443 MB) used the lowest memory, while SuperPoint (5305MB) and D2-Net (9521MB) would be too large to fit on the 4 GB GPU. Our method was the fastest with 14.8 ms for feature extraction, and does not rely on NMS to pick a subset of keypoints as SuperPoint (37.3 ms) and R2D2 (38.7 ms). D2-Net was the slowest with 477.4 ms, though it would likely be faster without multi-scale detection.

\subsection{Closed-loop: Lighting Change}

We repeated the path described in Section \ref{sec:vtr} 40 times with a 100\% autonomy rate. The repeats span lighting change from 3 a.m. until 10.30 p.m., see Figure \ref{fig:lighting-change-images}. All days had sunny weather with only occasional clouds. The robot repeated the path accurately for all conditions including driving in the dark with headlights. For this experiment only, we collected the path-tracking error measured with RTK-GPS. The root mean squared error (RMSE) across all repeats is 0.049 m with 0.070 m being the highest RMSE for an individual repeat. 
We provide box plots of the number of matched feature inliers in Figure \ref{fig:inlier_plots_light}. For every repeat, we get enough inliers to localize. Localization failure happens for frames with less than 6 inliers, which is the case for only three frames, see Table \ref{tab:loc_failure}. In this case, we rely on VO until localization recovers. As expected, we get the highest number of inliers in the middle of the day and the lowest number when it is dark. The dips in numbers during sunrise and sunset are caused by sun flares. Before sunrise there was fog, explaining the lower number of inliers compared to driving after dark at night.

\subsection{Closed-loop: Generalization to New Areas}

In the previous experiment, we showed that we can learn features based on data collected in 2016, teach a new path five years later and repeat it across large lighting change. We want to make sure that the network is not just memorizing known locations from the training data. This experiment tests generalization to areas and seasonal appearance not included in the training datasets. We train a network with data from both the In-The-Dark and Multiseason datasets. Both the on-road and off-road sections of the first path have one new unseen area, as shown in Figure \ref{fig:live_paths}, while the second and third paths are set in entirely new locations. The first path was repeated 26 times with a 100\% autonomy rate spanning lighting change from 4 a.m. until 9 p.m., see Figures \ref{fig:generalization-images}. Table \ref{tab:loc_failure} shows localization failures for a small part of the path driven during night-time. We provide box plots of the number of inliers for all repeats in Figure \ref{fig:inlier_plots} (a) and get the highest number of inliers during the day, some dips at sunrise and sunset due to sun flares, and the lowest inlier counts in the dark. In the same plot, we compare to localization using \ac{SURF} with colour-constant images \cite{Paton2016}. In eleven examples \ac{SURF} fails to localize at the beginning of the path. For the remaining repeats, the number of inliers is much lower than for the learned features. Since a large part of the path falls in an area with tree cover and other vegetation, we were unable to collect RTK GPS ground truth. In the new off-road area the robot swayed slightly at a few spots during the dark or strong sun flares, but quickly recovered and never drove off the path. In the new on-road area, for one repeat at 10:20, the robot turned too sharply and was off the path by roughly 1.5 meters, but recovered when exiting the turn.

We also present two plots in Figure \ref{fig:inlier_plots} (b) and (c) that compare the number of matched feature inliers when driving in the new areas versus the areas already seen in the training dataset. We divide the path into on-road and off-road sections since the on-road section generally gets more inliers. The new off-road area consistently gets fewer inliers than the area contained in the training data, but at the same time it gets enough inliers for driving. Moreover, the inlier numbers for the new area exhibit the same variation over time as those for the known area. In the new on-road area, the median number of inliers remains similar or higher compared to the known area. The reason is the new area is more similar to the rest of the path in the on-road case. Moreover, the new part of the path is less affected by sun flare at sunrise and sunset than the known on-road section, explaining why it has higher values at these times.

Finally, the second path collected in a field and the third path collected in a parking lot were both repeated 7 times with a 100\% autonomy rate from morning until dark. Figure \ref{fig:inlier_plots} (d) and (e) show a high number of inliers for both experiments, which means that the learned features generalized to these new places outside the training data. Localization failures only occur for the path in the field, see Table \ref{tab:loc_failure}.

\section{Conclusions and Future Work}

We have shown that we can use learned features for real-time autonomous path following under large lighting change and extend our path to new areas not seen in the data used to train the features. From data gathered in summer 2016 and from January to May 2017, we have trained networks to predict visual features that work reliably several years later. We used the existing sparse feature matcher in \ac{VTR}, but plan to implement dense matching in \ac{VTR} and use the learned scores in future work. We also aim to make the implementation more efficient so that we can drive faster. Finally, we plan a longer closed-loop experiment to test against seasonal change. 

\bibliographystyle{IEEEtran}
\bibliography{refs}

\begin{thebibliography}{10}
\providecommand{\url}[1]{#1}
\csname url@samestyle\endcsname
\providecommand{\newblock}{\relax}
\providecommand{\bibinfo}[2]{#2}
\providecommand{\BIBentrySTDinterwordspacing}{\spaceskip=0pt\relax}
\providecommand{\BIBentryALTinterwordstretchfactor}{4}
\providecommand{\BIBentryALTinterwordspacing}{\spaceskip=\fontdimen2\font plus
\BIBentryALTinterwordstretchfactor\fontdimen3\font minus
  \fontdimen4\font\relax}
\providecommand{\BIBforeignlanguage}[2]{{%
\expandafter\ifx\csname l@#1\endcsname\relax
\typeout{** WARNING: IEEEtran.bst: No hyphenation pattern has been}%
\typeout{** loaded for the language `#1'. Using the pattern for}%
\typeout{** the default language instead.}%
\else
\language=\csname l@#1\endcsname
\fi
#2}}
\providecommand{\BIBdecl}{\relax}
\BIBdecl

\bibitem{Paton2016}
M.~Paton, K.~MacTavish, M.~Warren, and T.~D. Barfoot, ``Bridging the appearance
  gap: Multi-experience localization for long-term visual teach and repeat,''
  in \emph{IROS}, 2016.

\bibitem{gridseth2020}
M.~Gridseth and T.~D. Barfoot, ``Deepmel: Compiling visual multi-experience
  localization into a deep neural network,'' in \emph{ICRA}, 2020.

\bibitem{Sattler2019}
T.~Sattler, Q.~Zhou, M.~Pollefeys, and L.~Leal-Taixe, ``Understanding the
  limitations of cnn-based absolute camera pose regression,'' in \emph{CVPR},
  2019.

\bibitem{piasco2019}
N.~Piasco, D.~Sidib{\'e}, V.~Gouet-Brunet, and C.~Demonceaux, ``Learning scene
  geometry for visual localization in challenging conditions,'' in \emph{ICRA},
  2019.

\bibitem{Stumberg2020a}
L.~von Stumberg, P.~Wenzel, Q.~Khan, and D.~Cremers, ``Gn-net: The gauss-newton
  loss for multi-weather relocalization,'' \emph{RAL}, 2020.

\bibitem{Stumberg2020b}
L.~Von~Stumberg, P.~Wenzel, N.~Yang, and D.~Cremers, ``Lm-reloc:
  Levenberg-marquardt based direct visual relocalization,'' in \emph{3DV},
  2020.

\bibitem{Kasper2020}
M.~Kasper, F.~Nobre, C.~Heckman, and N.~Keivan, ``Unsupervised metric
  relocalization using transform consistency loss,'' \emph{CoRR}, 2020.

\bibitem{spencer2020}
J.~Spencer, R.~Bowden, and S.~Hadfield, ``Same features, different day: Weakly
  supervised feature learning for seasonal invariance,'' in \emph{CVPR}, 2020.

\bibitem{Sarlin2021}
P.-E. Sarlin, A.~Unagar, M.~Larsson, H.~Germain, C.~Toft, V.~Larsson,
  M.~Pollefeys, V.~Lepetit, L.~Hammarstrand, F.~Kahl \emph{et~al.}, ``Back to
  the feature: Learning robust camera localization from pixels to pose,'' in
  \emph{CVPR}, 2021.

\bibitem{Gladkova2021}
M.~Gladkova, R.~Wang, N.~Zeller, and D.~Cremers, ``Tight integration of
  feature-based relocalization in monocular direct visual odometry,'' in
  \emph{ICRA}, 2021.

\bibitem{Sun2021}
L.~Sun, M.~Taher, C.~Wild, C.~Zhao, F.~Majer, Z.~Yan, T.~Krajnik, T.~Prescott,
  and T.~Duckett, ``Robust and long-term monocular teach-and-repeat navigation
  using a single-experience map,'' in \emph{IROS}, 2021.

\bibitem{chen2020survey}
C.~Chen, B.~Wang, C.~X. Lu, N.~Trigoni, and A.~Markham, ``A survey on deep
  learning for localization and mapping: Towards the age of spatial machine
  intelligence,'' \emph{arXiv preprint arXiv:2006.12567}, 2020.

\bibitem{Kendall2015}
A.~Kendall, M.~Grimes, and R.~Cipolla, ``Posenet: A convolutional network for
  real-time 6-dof camera relocalization,'' in \emph{ICCV}, 2015.

\bibitem{laskar2017}
Z.~Laskar, I.~Melekhov, S.~Kalia, and J.~Kannala, ``Camera relocalization by
  computing pairwise relative poses using convolutional neural network,'' in
  \emph{ICCV Workshops}, 2017.

\bibitem{Valada2018}
A.~Valada, N.~Radwan, and W.~Burgard, ``{Deep Auxiliary Learning for Visual
  Localization and Odometry},'' in \emph{ICRA}, 2018.

\bibitem{rosten2008}
E.~Rosten, R.~Porter, and T.~Drummond, ``Faster and better: A machine learning
  approach to corner detection,'' \emph{TPAMI}, 2008.

\bibitem{yi2016}
K.~M. Yi, E.~Trulls, V.~Lepetit, and P.~Fua, ``Lift: Learned invariant feature
  transform,'' in \emph{ECCV}, 2016.

\bibitem{ono2018}
Y.~Ono, E.~Trulls, P.~Fua, and K.~M. Yi, ``Lf-net: learning local features from
  images,'' in \emph{NeurIPS}, 2018, pp. 6234--6244.

\bibitem{detone2018a}
D.~DeTone, T.~Malisiewicz, and A.~Rabinovich, ``Superpoint: Self-supervised
  interest point detection and description,'' in \emph{CVPR Workshops}, 2018.

\bibitem{dusmanu2019}
M.~Dusmanu, I.~Rocco, T.~Pajdla, M.~Pollefeys, J.~Sivic, A.~Torii, and
  T.~Sattler, ``D2-net: A trainable cnn for joint description and detection of
  local features,'' in \emph{CVPR}, 2019.

\bibitem{revaud2019}
J.~Revaud, P.~Weinzaepfel, C.~R. de~Souza, and M.~Humenberger, ``{R2D2:}
  repeatable and reliable detector and descriptor,'' in \emph{NeurIPS}, 2019.

\bibitem{luo2020}
Z.~Luo, L.~Zhou, X.~Bai, H.~Chen, J.~Zhang, Y.~Yao, S.~Li, T.~Fang, and
  L.~Quan, ``Aslfeat: Learning local features of accurate shape and
  localization,'' in \emph{CVPR}, 2020.

\bibitem{sarlin2020}
P.-E. Sarlin, D.~DeTone, T.~Malisiewicz, and A.~Rabinovich, ``Superglue:
  Learning feature matching with graph neural networks,'' in \emph{CVPR}, 2020.

\bibitem{Wang2020}
Q.~Wang, X.~Zhou, B.~Hariharan, and N.~Snavely, ``Learning feature descriptors
  using camera pose supervision,'' in \emph{ECCV}, 2020.

\bibitem{Lv2019}
Z.~Lv, F.~Dellaert, J.~M. Rehg, and A.~Geiger, ``Taking a deeper look at the
  inverse compositional algorithm,'' in \emph{CVPR}, 2019.

\bibitem{Tang2019}
C.~Tang and P.~Tan, ``Ba-net: Dense bundle adjustment networks,'' in
  \emph{ICLR}, 2019.

\bibitem{Xu2020}
B.~Xu, A.~J. Davison, and S.~Leutenegger, ``Deep probabilistic feature-metric
  tracking,'' \emph{RAL}, 2020.

\bibitem{barnes2020}
D.~Barnes and I.~Posner, ``Under the radar: Learning to predict robust
  keypoints for odometry estimation and metric localisation in radar,'' in
  \emph{ICRA}, 2020.

\bibitem{Tang2020}
J.~Tang, R.~Ambrus, V.~Guizilini, S.~Pillai, H.~Kim, P.~Jensfelt, and
  A.~Gaidon, ``{Self-Supervised 3D Keypoint Learning for Ego-Motion
  Estimation},'' in \emph{CoRL}, 2020.

\bibitem{Hirschmuller2008}
H.~Hirschmuller, ``Stereo processing by semiglobal matching and mutual
  information,'' \emph{TPAMI}, 2008.

\bibitem{Barfoot2017}
T.~D. Barfoot, \emph{State Estimation for Robotics}.\hskip 1em plus 0.5em minus
  0.4em\relax Cambridge University Press, 2017.

\bibitem{christiansen2019}
P.~H. Christiansen, M.~F. Kragh, Y.~Brodskiy, and H.~Karstoft, ``Unsuperpoint:
  End-to-end unsupervised interest point detector and descriptor,'' \emph{arXiv
  preprint arXiv:1907.04011}, 2019.

\bibitem{Bay2006}
H.~Bay, T.~Tuytelaars, and L.~Van~Gool, ``Surf: Speeded up robust features,''
  in \emph{ECCV}, 2006.

\bibitem{Kingma2014}
D.~Kingma and J.~Ba, ``Adam: A method for stochastic optimization,''
  \emph{ICLR}, 2014.

\bibitem{Paton2018}
M.~Paton, K.~MacTavish, L.-P. Berczi, S.~K. van Es, and T.~D. Barfoot, ``I can
  see for miles and miles: An extended field test of visual teach and repeat
  2.0,'' in \emph{FSR}, 2018.

\end{thebibliography}

\end{document}